\documentclass[10pt,twocolumn,letterpaper]{article}

\usepackage[pagenumbers]{cvpr}
\definecolor{cvprblue}{rgb}{0.21,0.49,0.74}
\usepackage[pagebackref,breaklinks,colorlinks,allcolors=cvprblue]{hyperref}

\definecolor{calpolypomonagreen}{rgb}{0.12, 0.3, 0.17}
\usepackage[font=small, skip=4pt]{caption}
\usepackage{amsmath}
\usepackage{upgreek}
\usepackage{adjustbox}
\usepackage{xcolor}
\usepackage{color, colortbl}
\usepackage{booktabs}
\usepackage{arydshln}
\usepackage{multirow}
\usepackage{algorithm}
\usepackage{algpseudocode}
\usepackage{graphicx}
\usepackage{tabularx}
\usepackage{animate}
\usepackage{amsfonts}

\definecolor{theme}{RGB}{190,184,220}
\colorlet{theme100}{theme!100}
\colorlet{theme80}{theme!80}
\colorlet{theme60}{theme!60}
\colorlet{theme40}{theme!40}
\colorlet{theme20}{theme!20}
\colorlet{1st}{theme!100}
\colorlet{2nd}{theme!50}
\colorlet{3rd}{theme!20}
\colorlet{4th}{theme!0}
\colorlet{5th}{theme!0}

\title{ChronosObserver: Taming 4D World with Hyperspace Diffusion Sampling}

\author{Qisen~Wang\;\; Yifan~Zhao$^*$\;\; Peisen~Shen\;\; Jialu~Li\;\; Jia~Li$^*$\\
State Key Laboratory of Virtual Reality Technology and Systems, SCSE, Beihang University\\
{\tt\small \{wangqisen, zhaoyf, psshen, ljl16, jiali\}@buaa.edu.cn}
}

\begin{document}
\twocolumn[{%
\renewcommand\twocolumn[1][]{#1}%
\maketitle
\begin{center}
    \centering
    \captionsetup{type=figure}
    \vspace{-1.5em}
    \includegraphics[width=1\textwidth]{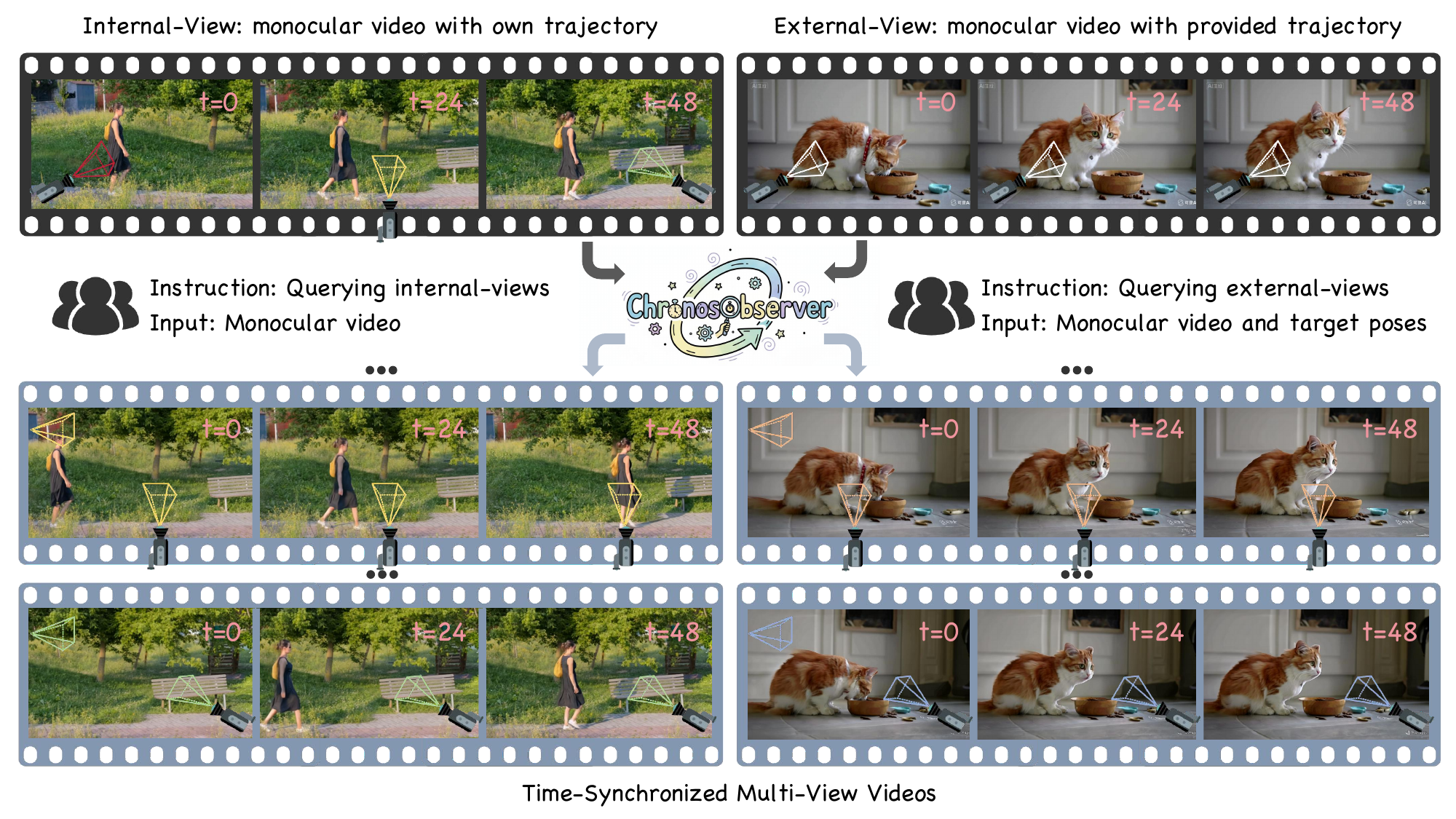}
    \vspace{-1.5em}
    \captionof{figure}{\textbf{ChronosObserver results of temporal-synchronized multi-view videos from one single monocular video}. The project page is \url{https://icvteam.github.io/ChronosObserver.html}.}
    \label{fig:teaser}
\end{center}%
}]
\def\thefootnote{}\footnotetext{$^*$ Correspondence should be addressed to Yifan Zhao and Jia Li. Website: \url{https://cvteam.buaa.edu.cn}.}
\begin{abstract}
Although prevailing camera-controlled video generation models can produce cinematic results, lifting them directly to the generation of 3D-consistent and high-fidelity \textbf{time-synchronized multi-view videos} remains challenging, which is a pivotal capability for taming 4D worlds. Some works resort to data augmentation or test-time optimization, but these strategies are constrained by limited model generalization and scalability issues. To this end, we propose ChronosObserver, a training-free method including World State Hyperspace to represent the spatiotemporal constraints of a 4D world scene, and Hyperspace Guided Sampling to synchronize the diffusion sampling trajectories of multiple views using the hyperspace. Experimental results demonstrate that our method achieves high-fidelity and 3D-consistent time-synchronized multi-view videos generation \textbf{without training or fine-tuning} for diffusion models.
\end{abstract}    
\section{Introduction}
\label{sec:intro}

\begin{figure}[t]
    \centering
    \includegraphics[width=\columnwidth]{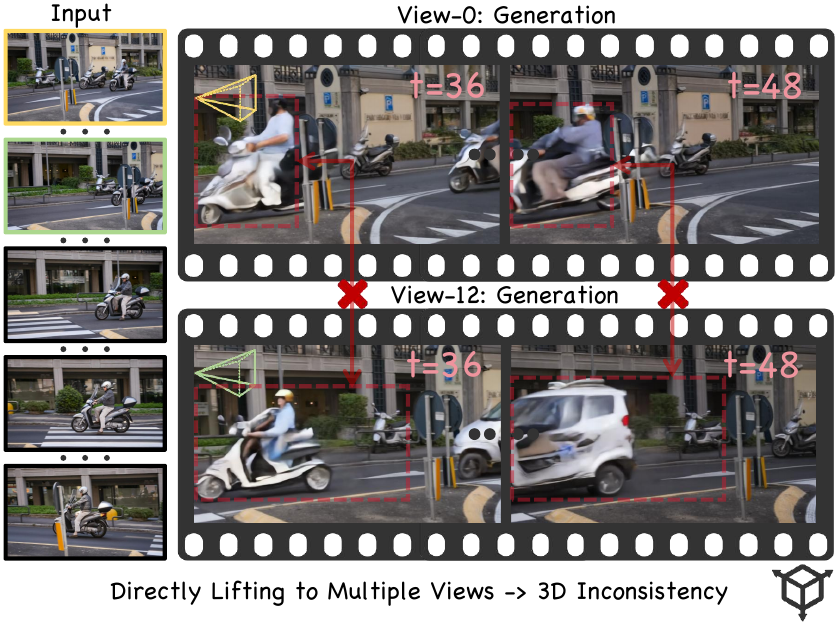}
    \caption{\textbf{Motivation}. Directly lifting a camera-controlled video generation method~\cite{trajectorycrafter} to generate multi-view videos leads to 3D inconsistencies across different viewpoints at the same timestamp.}
    \label{fig:motivation}
    \vspace{-1.5em}
\end{figure}

Video generation~\cite{video_generation_survey_1, video_generation_survey_2} has achieved remarkable progress, particularly with the development of camera-controlled video generation~\cite{cameractrl, recammaster, trajectorycrafter} that can produce cinematic visual results of view-controlling over specified camera trajectories. However, a critical challenge toward the 4D world that remains largely unaddressed is the generation of 3D-consistent and high-fidelity time-synchronized multi-view videos.
As shown in \cref{fig:motivation}, a straightforward extension of the existing camera-controlled video generation method~\cite{trajectorycrafter} to multi-view synthesis frequently produces \textbf{\textit{3D inconsistencies}}. This issue is rooted in independent sampling trajectories across viewpoints, serving as the sparse conditioning constraints, which undermine \textbf{\textit{weakly-coupled conditional distributions}} for different viewpoints, leading to the failure to enforce a \textbf{\textit{unified 4D scene}}.
Although prevailing methods attempt to address this challenge using data augmentation~\cite{wu2025cat4d, sun2025dimensionx} or per-scene test-time optimization~\cite{rav}, these remain constrained by inherent defects, including \textit{limited model generalization} due to insufficient multi-view data diversity and \textit{scalability issues} arising from crucial computational requirements.

The 3D inconsistencies shown in \cref{fig:motivation} underscore a fundamental multi-view disjointedness in generation when naively extending existing methods. But a critical insight emerges from the observation that these models demonstrate a compelling capacity to produce geometrically consistent and high-fidelity content for each individual camera trajectory. This confirms that existing camera-controlled generation models encapsulate the inherent spatiotemporal coherence. Therefore, the central problem lies not in model ability but in achieving \textbf{\textit{cross-view synchronization}}, as the current paradigm lacks a mechanism to enforce coupling across the sampling trajectories of conditional distributions for different viewpoints. This situation shifts the core challenge toward determining \textbf{\textit{how to structure the cross-view sampling trajectories}} to induce a unified 4D scene. Under the constraint of a single monocular video input, addressing this question necessitates the design of \textbf{\textit{explicit spatiotemporal constraints}} that can synchronize the generative process across views, thereby merging independent sampling results into a coherent 4D scene.

To this end, we propose \textbf{ChronosObserver}, a \textit{training-free} method for the challenge of \textit{time-synchronized multi-view videos generation}. 
First, we propose \textbf{World State Hyperspace}, formulated as an \textit{incremental representation} that models the unified spatiotemporal states of a 4D scene to provide compatible constraints throughout generation.
Furthermore, we propose \textbf{Hyperspace Guided Sampling}, which leverages spatiotemporal constraints within the hyperspace to \textit{control the diffusion generation process} 
by integrating these constraints within the latent space during denoising.
For \textbf{\textit{any view and any time}}, our proposed method achieves 3D-consistent time-synchronized multi-view videos generation.

Overall, our contributions can be summarized as
\begin{enumerate}
    \item We propose \textbf{ChronosObserver}, a training-free method for generating time-synchronized multi-view videos.
    \item We propose \textbf{World State Hyperspace}, an incremental representation that encapsulates the spatiotemporal constraints of a 4D world scene.
    \item We propose \textbf{Hyperspace Guided Sampling} for guiding the diffusion sampling using the hyperspace to achieve 3D-consistent and high-fidelity generation results.
\end{enumerate}

\section{Related Works}
\label{sec:related_works}

\textbf{Video Generation.}
Video generation models~\cite{video_generation_survey_1, video_generation_survey_2} have advanced rapidly, with mainstream approaches including diffusion-based~\cite{diffusion2015, ddpm, sde, stable_diffusion} models~\citep{he2022latent, hunyuanvideo} as well as recently emerging autoregressive models~\cite{yin2025slow, dengautoregressive}. These models demonstrate a strong capacity for generating high-fidelity videos with realistic spatiotemporal visual information.
Leveraging the capabilities of foundation models~\cite{svd, cogvideox, wan}, researchers have begun to explore controllable video generation, \eg, motion-controlled~\cite{cao2025uni3c, ma2025follow} and camera-controlled generation~\cite{cameractrl, recammaster, gen3c, wu2025video}, leading to increasingly realistic and controllable video content.
A key insight from recent studies is that video generation models exhibit an emergent understanding of real-world spatial relationships~\cite{sora, cosmos} when scaled sufficiently in both parameters and training data. This has motivated growing interest in camera-controlled generation techniques~\cite{cameractrl, trajectoryattention, trajectorycrafter}. As a result, supplementing and extrapolating 4D visual information through generative approaches has become a challenging yet promising research direction.

\begin{figure*}[t]
    \centering
    \includegraphics[width=\textwidth]{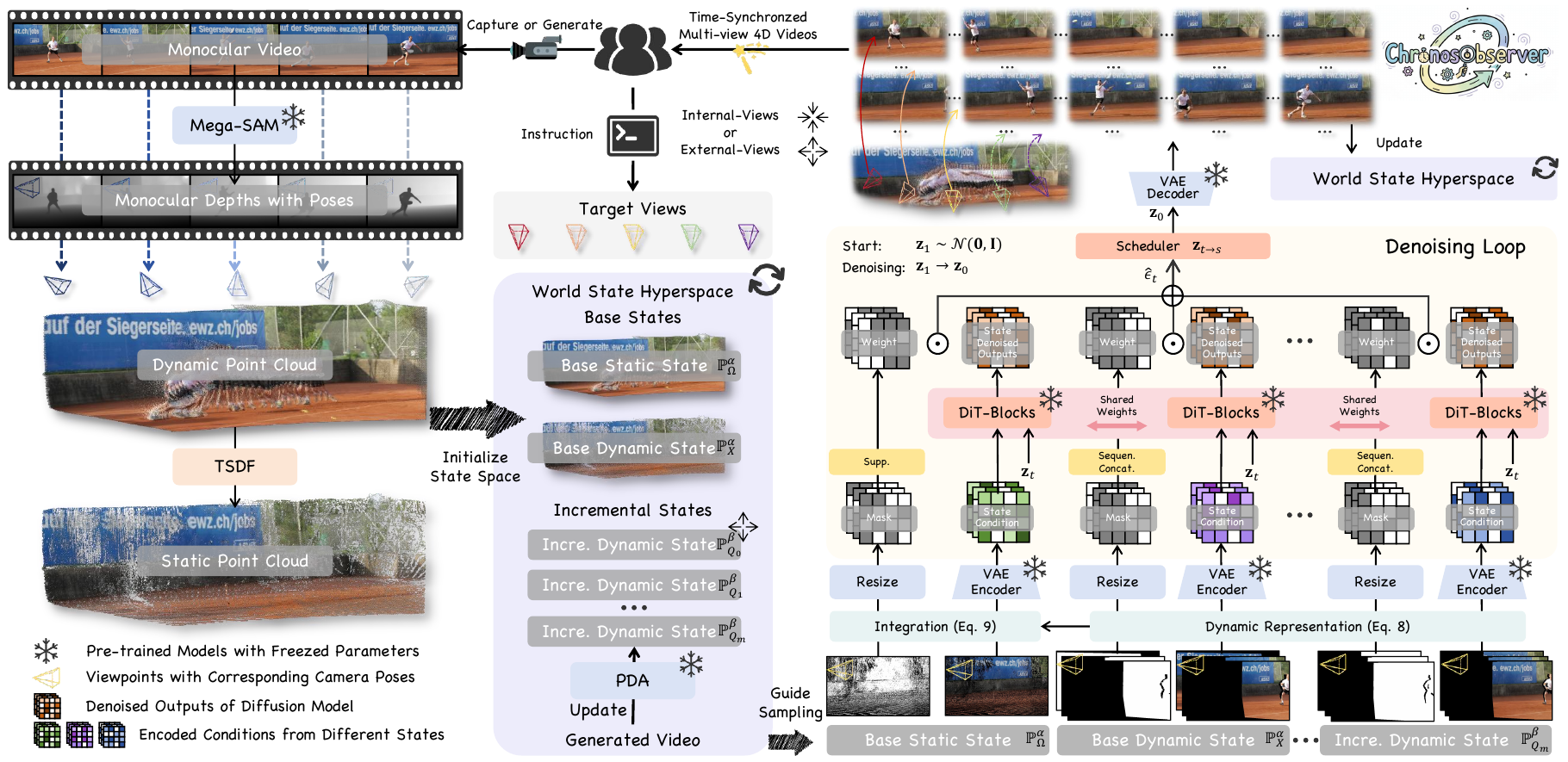}
    \caption{\textbf{ChronosObserver Pipeline} starts from the monocular input video and generates time-synchronized multi-view videos. ChronosObserver incrementally constructs the World State Hyperspace and utilizes it for the Hyperspace Guided Sampling.}
    \label{fig:pipeline}
    \vspace{-1.5em}
\end{figure*}
\textbf{Camera-controlled Video Generation.}
While existing novel view synthesis methods~\cite{nerf, 3dgs} excel for dense inputs, they often struggle with sparse or monocular inputs due to insufficient visual supervision. This persistent limitation is driving a shift towards generative approaches, namely camera-controlled video generation models~\cite{you2024nvs, houtraining, wang2024motionctrl, bahmanivd3d, ma2025you, xiao20243dtrajmaster, wang2025cinemaster}, which leverage diffusion priors to generate visual content at target poses from a single image or video conditioned on camera trajectories.
Existing methods for camera control follow two main paradigms.
Some approaches~\cite{cameractrl, cameractrl2, recammaster} implicitly encode the target camera poses as network input, but face issues like limited control over pose scale and absence of explicit visual guidance.
In contrast, ViewCrafter~\cite{viewcrafter} first introduces point cloud representations to provide explicit pose guidance.
Subsequent works~\cite{trajectorycrafter, ex4d, rav} extract dynamic point clouds via monocular depth estimation and reproject them to target poses for reliable control.
However, these methods typically operate on a single dynamic camera trajectory to achieve cinematic effects. A critical challenge lies in generating time-synchronized multi-view videos from sparse visual clues. While essential for taming the 4D world, this task represents a key open problem for existing camera-controlled video generation models.

\textbf{Time-synchronized Multi-view Video Generation.} As previously motioned, time-synchronized multi-view videos generation~\cite{xucavia, wu2025cat4d, baisyncammaster, kuang2024collaborative, van2024generative, wang20254real, sun2025dimensionx} represents a critical step toward constructing the 4D world.
However, due to challenges in data acquisition, there remains a severe scarcity of multi-view dynamic scene datasets.
Thus, prevailing works~\cite{wu2025cat4d, sun2025dimensionx} attempt to augment data, like leveraging real-world multi-view static scene data, general video datasets, and synthetic 4D data for training. Despite these efforts, the fundamental issue of data scarcity persists and remains inadequately addressed.
To this end, Reangle-A-Video~\cite{rav} employs the pre-trained video generation model combined with test-time fine-tuning. While this strategy alleviates data dependency, it compromises high inference computational resources.
In this work, we propose a training-free framework to efficiently and effectively generate high-quality time-synchronized multi-view 4D videos without large-scale datasets for fine-tuning and test-time optimization for diffusion models.

\section{Method}
\label{sec:method}

Towards the challenge of 3D inconsistencies over viewpoints identified in \cref{sec:intro}, we propose ChronosObserver, a training-free method that enforces a unified 4D scene by structuring the diffusion sampling process with explicit spatiotemporal constraints.
As shown in \cref{fig:pipeline}, we propose World State Hyperspace to incrementally maintain consistent scene representation and Hyperspace Guided Sampling to synchronize the generation process for different views.

\textbf{Formulation and Initialization.} 
Given the input monocular video $\mathcal{I}$ with $T$ frames $\mathcal{I} = \{I_0, I_1, ..., I_{T-1}\}$, we first utilize Mega-SAM~\cite{megasam} to predict its depths $\mathcal{D} = \{D_0, D_1, ..., D_{T-1}\}$, poses $\mathcal{P} = \{P_0, P_1, ..., P_{T-1}\}$, and the intrinsic $K$.
In this work, we consider two scenarios for the target poses $\mathcal{Q}=\{Q_m\}_{m=1,...,M}$: \textit{internal-view}, where the target poses are set to the internal poses of the monocular video; and \textit{external-view}, where the target poses are set to the external poses specified by the user.
Meanwhile, we construct the dynamic trajectory as $Q_0 = Q_{1 \rightarrow M}$ for providing an additional auxiliary reference. For the internal-view scenario, $Q_0$ corresponds to the input video and can be ignored.
We aim to generate time-synchronized multi-view 4D videos $\mathcal{I}^{\mathcal{Q}} = \{\mathcal{I}^{Q_m}\}$ from the given target poses $\mathcal{Q}$.

\textbf{World State Hyperspace.} 
Existing camera-controlled video diffusion models can individually generate the video of the target pose $Q_m$ from the source video $\mathcal{I}_*$ and its depths $\mathcal{D}_*$, poses $\mathcal{P}_*$, with the guidance of the view-warped visual results of the input video. Specifically, we have the conditional distribution as
\begin{equation}
    \mathcal{I}_*^{Q_m} \sim p_{\theta}(\mathcal{I}_*^{Q_m} | \mathcal{I}^{Q_m}_{*;r}, \mathcal{M}^{Q_m}_{*;r}, Q_m), 
\end{equation}
where condition $\mathcal{I}_*$ is omitted for simplification, and the same applies below. $\mathcal{I}^{Q_m}_{*;r}, \mathcal{M}^{Q_m}_{*;r}$ are obtained from the view-warping of each frame $i$ with the warping function $\Psi$ and the projection function $\psi$, which can be represented as
\begin{equation}
\label{eq:warp}
    \begin{aligned}
        &\mathcal{I}^{Q_m}_{*;r}|_i, \mathcal{M}^{Q_m}_{*;r}|_i = \Psi(\mathcal{I}_*, \mathcal{D}_*, K, \mathcal{P}_*, Q_m, i), \\
        &= \psi(\psi^{-1}(\mathcal{I}_*|_i, \mathcal{D}_*|_i, K, \mathcal{P}_*|_i), K, Q_m).
    \end{aligned}
\end{equation}
Here, we denote the unprojected points
\begin{equation}
\label{eq:unproj_points}
    \mathbb{P}_* = \psi^{-1}(\mathcal{I}_*|_i, \mathcal{D}_*|_i, K, \mathcal{P}_*|_i) |_{i=0,...,T-1}
\end{equation}
from the source video $\mathcal{I}_*$ as the world state, which serves as the constraint of guiding the trajectory of diffusion sampling. So, the conditional distribution can also be represented as
\begin{equation}
    \mathcal{I}_*^{Q_m} \sim p_{\theta}(\mathcal{I}_*^{Q_m} | \mathbb{P}_*, Q_m), 
\end{equation}
where $\mathcal{I}_*^{Q_m}$ is conditioned by the source video and the world state with the target pose.
The generation inverts the diffusion process in the latent space, starting from noise $\mathbf{z}_1 \sim \mathcal{N}(\mathbf{0}, \mathbf{I})$ and iteratively denoising it to obtain $\mathbf{z}_0$, which is decoded into $\mathcal{I}_*^{Q_m}$. The denoising step from a noisy latent $\mathbf{z}_t$ to $\mathbf{z}_s$ ($1 \geq t > s \geq 0$) can be represented as
\begin{equation}
\label{eq:denoise_scheduler}
    \mathbf{z}_{t \rightarrow s} = \text{Scheduler}(\mathbf{z}_t, \hat{\epsilon}_t=\epsilon_{\theta}(\mathbf{z}_t, \mathbf{x}_{*;r}^{Q_m}, t), t, s),
\end{equation}
where $\mathbf{x}_{*;r}^{Q_m}$ is the encoded condition of $\mathcal{I}^{Q_m}_{*;r}, \mathcal{M}^{Q_m}_{*;r}$, and $\epsilon_{\theta}$ is the parameterized noise predictor.

However, we can notice that the produced videos from the existing camera-controlled video diffusion models lack the 3D constraints across different views.
Therefore, we propose the concept of the world state hyperspace $\zeta$, which includes the existing constraints for controlling the diffusion sampling.
Specifically, the initialized space $\{\mathbb{P}_{X}^{\alpha}, \mathbb{P}_{\Omega}^{\alpha}\}$ includes the dynamic points $\mathbb{P}_{X}^{\alpha}$ and the static points $\mathbb{P}_{\Omega}^{\alpha}$ of the Truncated Signed Distance Function (TSDF)~\cite{tsdf} from the input $\mathcal{I}$ and the additional auxiliary reference $\mathcal{I}^{Q_0}$ with only using $\mathbb{P}_{X}^{\alpha}$, which are denoted as the \textbf{Base State Hyperspace} $\zeta^{\alpha}=\{ \mathbb{P}_{X}^{\alpha}, \mathbb{P}_{\Omega}^{\alpha} \}$.

Similar to the idea of mathematical induction, we temporarily do not consider how to efficiently supplement and use the world states. Instead, we start from the base state hyperspace with 3D consistency, and aim to supplement the generated 3D-consistent \textbf{Incremental State Hyperspace} $\zeta^{\beta}_*$, where $*$ represents the current status.
We assume that we have already generated the video $\mathcal{I}^{Q_m}$ with bullet-time 3D consistency of the target pose $Q_m$ and supplement its corresponding state $\mathbb{P}_{Q_m}^{\beta}$ to the incremental state hyperspace $\zeta^{\beta}_m = \zeta^{\beta}_{m-1} \cup \{\mathbb{P}_{Q_m}^{\beta}\}$, which follows \cref{eq:unproj_points} and can be represented as
\begin{equation}
\label{eq:unproj_points_qm}
    \begin{aligned}
        \mathbb{P}_{Q_m}^{\beta} &= \psi^{-1}(\mathcal{I}^{Q_m}|_i, \mathcal{D}^{Q_m}|_i, K, Q_m) |_{i=0,...,T-1}, \\
        \mathcal{D}^{Q_m} &= \text{PDA}(\mathcal{I}^{Q_m}, \mathcal{D}^{\mathcal{P}\rightarrow\mathcal{Q}_m}(\mathcal{D}, \mathcal{P}, \mathcal{Q}_m)),
    \end{aligned}
\end{equation}
where $\mathcal{D}^{Q_m}$ is predicted by Prior Depth Anything (PDA)~\cite{pda} using the generated video $\mathcal{I}^{Q_m}$ and the warped depth $\mathcal{D}^{\mathcal{P}\rightarrow\mathcal{Q}_m}$ from the depth of the input video. Note that we skip supplementing the states where there are no valid pixels of warped depth from the input frame.
The supplemented states are denoted as the incremental state hyperspace.
Note that $\zeta^{\beta}_0$ is initialized by the generated $\{\mathbb{P}_{Q_0}^{\beta}\}$ with related to $\mathcal{I}^{Q_0}$. 
To this end, we can notice that if we can find a way to constrain the sampling trajectory of the camera-controlled video diffusion $p_{\theta}$ using the world state hyperspace to generate the incremental video of the target pose $Q_m$, we can generate the time-synchronized multi-view 4D videos $\mathcal{I}^{\mathcal{Q}} = \{\mathcal{I}^{Q_m}\}$ which are what we want.

\begin{figure}[t]
    \centering
    \includegraphics[width=\columnwidth]{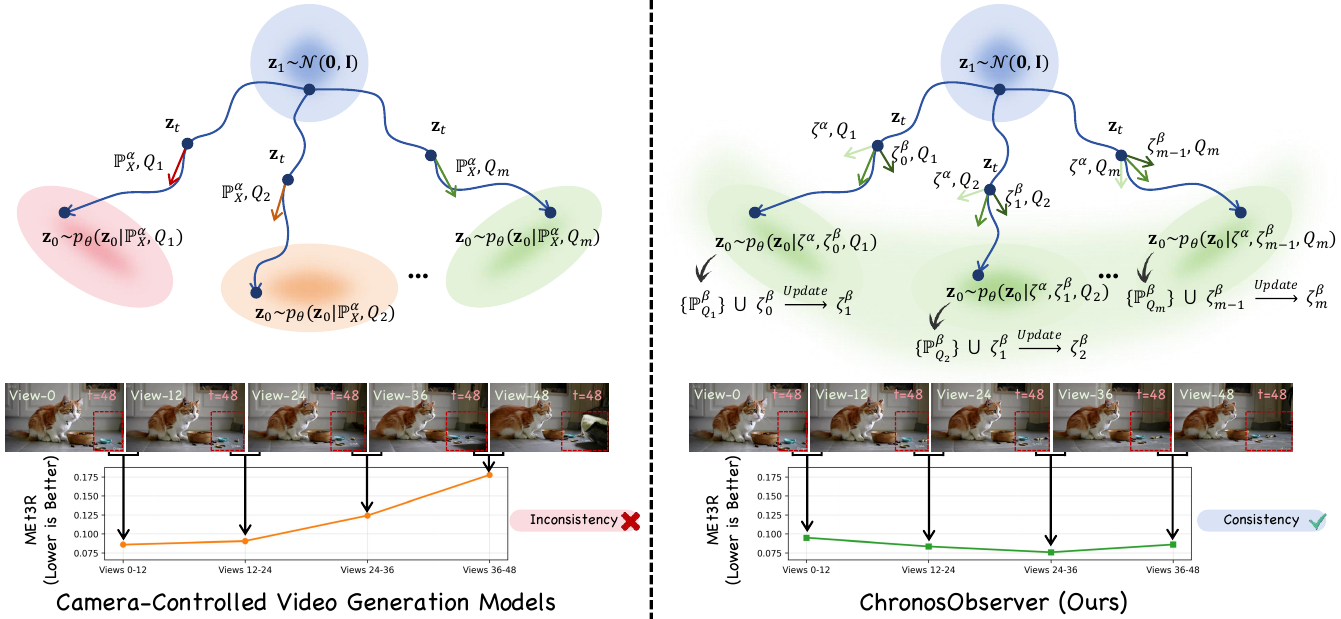}
    \caption{\textbf{Intuitive Illustration} of ChronosObserver compared to camera-controlled video generation methods.}
    \label{fig:diffusion_sampling}
    \vspace{-1.5em}
\end{figure}
\begin{table*}[t]
\centering
\caption{\textbf{Quantitative comparison results}.}
\setlength{\tabcolsep}{1.5mm}{
\begin{tabular}{c||c|cccccc}
\hline
\cellcolor{gray!10} & \cellcolor{gray!10}\textbf{3D Consistency} & \multicolumn{6}{c}{\cellcolor{gray!10}\textbf{VBench}} \\
\cline{2-8}
\cellcolor{gray!10}\multirow{-2}{*}{\textbf{Methods}} & \cellcolor{gray!10}MEt3R$\downarrow$ & \cellcolor{gray!10}Subject$\uparrow$ & \cellcolor{gray!10}Background$\uparrow$ & \cellcolor{gray!10}Motion$\uparrow$ & \cellcolor{gray!10}Temporal$\uparrow$ & \cellcolor{gray!10}Aesthetic$\uparrow$ & \cellcolor{gray!10}Imaging$\uparrow$ \\
\hline\hline
ViewCrafter \cite{viewcrafter} & \cellcolor{5th}0.3086 & \cellcolor{3rd}0.9038 & \cellcolor{3rd}0.9403 & \cellcolor{4th}0.9792 & \cellcolor{2nd}0.9829 & \cellcolor{4th}0.4907 & \cellcolor{3rd}0.5909 \\
Reangle-A-Video \cite{rav} & \cellcolor{3rd}0.2342 & \cellcolor{4th}0.8408 & \cellcolor{4th}0.9070 & \cellcolor{5th}0.9788 & \cellcolor{5th}0.9583 & \cellcolor{3rd}0.5303 & \cellcolor{4th}0.5595 \\
EX-4D \cite{ex4d} & \cellcolor{4th}0.2514 & \cellcolor{5th}0.8267 & \cellcolor{5th}0.8935 & \cellcolor{3rd}0.9881 & \cellcolor{4th}0.9769 & \cellcolor{5th}0.4873 & \cellcolor{5th}0.5546 \\
TrajectoryCrafter \cite{trajectorycrafter} & \cellcolor{2nd}0.1930 & \cellcolor{2nd}0.9238 & \cellcolor{2nd}0.9475 & \cellcolor{2nd}0.9886 & \cellcolor{3rd}0.9773 & \cellcolor{1st}\textbf{0.5660} & \cellcolor{2nd}0.6149 \\
ChronosObserver (Ours) & \cellcolor{1st}\textbf{0.1635} & \cellcolor{1st}\textbf{0.9365} & \cellcolor{1st}\textbf{0.9582} & \cellcolor{1st}\textbf{0.9918} & \cellcolor{1st}\textbf{0.9834} & \cellcolor{2nd}0.5598 & \cellcolor{1st}\textbf{0.6332} \\
\hline
\end{tabular}
}
\label{tab:comp_sota}
\vspace{-1em}
\end{table*}
\textbf{Hyperspace Guided Sampling.} 
As demonstrated above, with the world state hyperspace and the target pose $Q_m$, we need to achieve 3D-consistent sampling on the target pose $Q_m$ through guiding the sampling trajectory of the video diffusion model using the state space. 
Specifically, we can obtain the projected rendered frames and masks for each state $\mathbb{P}_*$ in current $\zeta_{m-1} = \zeta^{\alpha} \cup \zeta^{\beta}_{m-1}$ following \cref{eq:warp}, which can be represented as
\begin{equation}
    \mathcal{I}_{*;r}^{Q_m}, \mathcal{M}_{*;r}^{Q_m} = \psi(\mathbb{P}_*, K, Q_m).
\end{equation}
For ease of description, we denote $\mathcal{I}_{\omega;r}^{Q_m}, \mathcal{M}_{\omega;r}^{Q_m}$ for base static state, $\mathcal{I}_{\chi;r}^{Q_m}, \mathcal{M}_{\chi;r}^{Q_m}$ for base dynamic state, and $\mathcal{I}_{j;r}^{Q_m}, \mathcal{M}_{j;r}^{Q_m}$ for $j$-th incremental dynamic state.
Then, we compute the existing dynamic representation by taking the union of the dynamic states, which can be represented as
\begin{equation}
\label{eq:accum_render}
    \begin{aligned}
        \mathcal{M}_{\mu;\ast}^{Q_m} &= \mathcal{M}_{\chi;r}^{Q_m} \lor (\bigvee_{j=0}^{\ast-1} \mathcal{M}_{j;r}^{Q_m}), \\
        \mathcal{I}_{\mu;\ast}^{Q_m} &= \mathcal{M}_{\chi;r}^{Q_m} \mathcal{I}_{\chi;r}^{Q_m} + \sum_{j=0}^{\ast-1} (\lnot \mathcal{M}_{\mu;j}^{Q_m} \land \mathcal{M}_{j;r}^{Q_m}) \mathcal{I}_{j;r}^{Q_m},
    \end{aligned}
\end{equation}
and $\mathcal{I}_{\mu;m}^{Q_m}, \mathcal{M}_{\mu;m}^{Q_m}$ are the current accumulated dynamic representation. We further integrate the dynamic representation into the static representation to aid subsequent diffusion sampling, which can be written as
\begin{equation}
\label{eq:integrate_static_dynamic}
    \begin{aligned}
        \mathcal{M}_{\omega,\mu;r} &= \mathcal{M}_{\mu;m}^{Q_m} \lor \mathcal{M}_{\omega;r}^{Q_m}, \\
        \mathcal{I}_{\omega,\mu;r}^{Q_m} &= \mathcal{M}_{\mu;m}^{Q_m} \mathcal{I}_{\mu;m}^{Q_m} + (\lnot \mathcal{M}_{\mu;m}^{Q_m} \land \mathcal{M}_{\omega;r}^{Q_m}) \mathcal{I}_{\omega;r}^{Q_m}.
    \end{aligned}
\end{equation}
After encoding the conditions rendered by the above states, we obtain the state representations in the latent space, which are denoted by $\mathbf{x}_{\omega,\mu;r}^{Q_m}$ for $\mathcal{I}_{\omega,\mu;r}^{Q_m}, \mathcal{M}_{\omega,\mu;r}^{Q_m}$, $\mathbf{x}_{\chi;r}^{Q_m}$ for $\mathcal{I}_{\chi;r}^{Q_m}, \mathcal{M}_{\chi;r}^{Q_m}$, and $\mathbf{x}_{j;r}^{Q_m}$ for $\mathcal{I}_{j;r}^{Q_m}, \mathcal{M}_{j;r}^{Q_m}$.
Besides, we construct latent weights $\mathbf{w}_{\chi;r}^{Q_m}, \mathbf{w}_{j;r}^{Q_m}$ through sequential mask concatenation, and employ $\mathbf{w}_{\omega,\mu;r}^{Q_m}$ to handle residual uncovered areas, which ensures complete spatial coverage while maintaining normalized weight distribution across all states.
Given the pre-trained noise predictor of the camera-controlled video generation model $\epsilon_{\theta}(\cdot)$, we fuse the prediction results of state representations at each step of the denoising loop to perform multi-state representation constraint sampling, which can be represented as
\begin{equation}
    \begin{aligned}
        \hat{\epsilon}_t &= \mathbf{w}_{\omega,\mu;r}^{Q_m}  \epsilon_{\theta}(\mathbf{z}_t, \mathbf{x}_{\omega,\mu;r}^{Q_m}, t) + \mathbf{w}_{\chi;r}^{Q_m} \epsilon_{\theta}(\mathbf{z}_t, \mathbf{x}_{\chi;r}^{Q_m}, t) \\
        &+ \sum_{j=0}^{m-1} \mathbf{w}_{j;r}^{Q_m} \epsilon_{\theta}(\mathbf{z}_t, \mathbf{x}_{j;r}^{Q_m}, t).
    \end{aligned}
\end{equation}
We can further get the denoised output $\mathbf{z}_{t \rightarrow s}$ similar to the standard diffusion generation process shown in \cref{eq:denoise_scheduler}.

\textbf{Intuitive Illustration.}
We further elaborate on the effectiveness of the proposed world state constraints sampling from the perspective of the conditional distribution of the video diffusion model.
As shown in the left part of \cref{fig:diffusion_sampling}, the previous camera-controlled video generation models only utilize one single state $\mathbb{P}_{X}^{\alpha}$ from the input monocular video, for guiding the diffusion sampling, which can be represented as
\begin{equation}
    \mathcal{I}^{Q_m} \sim p_{\theta}(\mathcal{I}^{Q_m} | \mathbb{P}_{X}^{\alpha}, Q_m),\;\; \mathbb{P}_{X}^{\alpha} \in \zeta^{\alpha}.
\end{equation}
Since the projection results obtained by state $\mathbb{P}_{X}^{\alpha}$ at different target poses are different, and the sampling processes are independent of each other, the condition distributions corresponding to different target poses are unsynchronized, which in turn leads to 3D inconsistency in the generated time-synchronized multi-view videos.
Conversely, our proposed method constructs the world state hyperspace as described above, and then fuses the conditional distributions corresponding to these world states during the video generation process for each target pose, thereby obtaining the sampling results constrained by the hyperspace, as shown in the right part of \cref{fig:diffusion_sampling}. Furthermore, due to the self-iterative property of the hyperspace, our time-synchronized multi-view video sampling process is autoregressive, namely
\begin{equation}
    \begin{aligned}
        &\mathcal{I}^{Q_m} \sim p_{\theta}(\mathcal{I}^{Q_m} | \zeta^{\alpha}, \zeta^{\beta}_{m-1}, Q_m) \\
        &= p_{\theta}(\mathcal{I}^{Q_m} | \zeta^{\alpha}, \zeta^{\beta}_{m-1}=\zeta^{\beta}_{m-2}\cup\{\mathbb{P}_{Q_{m-1}}^{\beta}\}, Q_m),
    \end{aligned}
\end{equation}
where $\zeta^{\beta}_{m-2}$ is related to $\mathcal{I}^{Q_0}, ..., \mathcal{I}^{Q_{m-2}}$, and $\mathbb{P}_{Q_{m-1}}^{\beta}$ is related to $\mathcal{I}^{Q_{m-1}}$.
Since we construct the incremental state hyperspace corresponding to the identical 4D scene, and utilize the hyperspace to jointly constrain the sampling trajectory of video generation, we can achieve 3D-consistent time-synchronized multi-view 4D videos.
\section{Experiments}
\label{sec:exp}

\subsection{Experimental Details}
\label{sec:exp_details}

\textbf{Dataset.}
We evaluate our method on our collected 30 monocular videos, which include 20 monocular videos in the wild from DAVIS~\cite{davis}, and 10 monocular videos generated by the Kling video generation model~\cite{kling}. 
Each video contains 49 frames with a resolution of $384\times672$. For original videos that are too long, we will truncate and extract frames.
For the monocular videos with wide camera trajectories filmed in the wild, we utilize the fixed internal perspective of the videos for inference, which are called Interval-View Scenes.
For the monocular videos with relatively static camera trajectories generated by the video generation model, we provide the fixed external perspectives for inference, which are called External-View Scenes.
Besides, top-10 cases are extracted through the MEt3R scores of TrajectoryCrafter~\cite{trajectorycrafter}, which are called Hard Scenes.
See more dataset details in the Appendix.

\begin{figure*}[t]
    \centering
    \includegraphics[width=\textwidth]{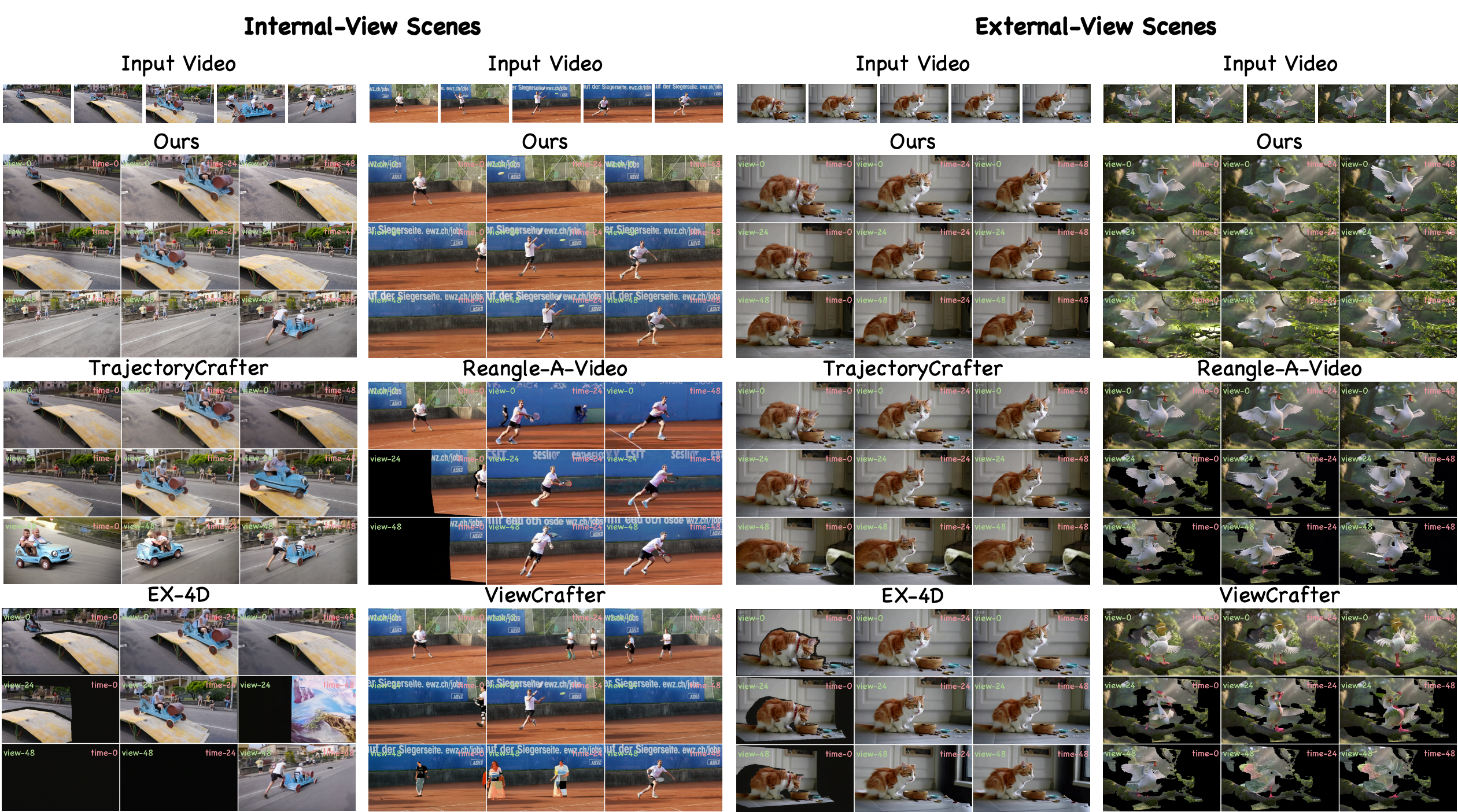}
    \caption{\textbf{Qualitative comparisons with other methods} (TrajectoryCrafter~\cite{trajectorycrafter}, EX-4D~\cite{ex4d}, Reangle-A-Video~\cite{rav}, ViewCrafter~\cite{viewcrafter}).}
    \label{fig:comp_sotas_1}
    \vspace{-1.0em}
\end{figure*}
\textbf{Implementation.}
We utilize TrajectoryCrafter~\cite{trajectorycrafter} as the video generation model of our method.
The diffusion sampling step is set to $30$, and the Classifier-Free Guidance (CFG) scale is set to $6.0$, which is consistent to TrajectoryCrafter.
For both Interval-View and External-View Scenes, we extract five perspectives of $[0, 12, 24, 36, 48]$ as our target poses.
The code is based on PyTorch and CUDA 12.1.
Experiments are conducted on 48GB 4090.
The computational time cost is around 10 minutes per target view.
See more implementation details in the Appendix.

\textbf{Metrics.}
We utilize VBench~\cite{vbench} to evaluate the quality of generated videos, including Subject Consistency, Background Consistency, Motion Smoothness, Temporal Flickering, Aesthetic Quality, and Imaging Quality.
Furthermore, we utilize the multi-view 3D consistency metric MEt3R~\cite{met3r} to evaluate the 3D consistency of generated videos on different sequential pair-wise target views at the same timestamp. 
See more details of metrics in the Appendix.

\textbf{Baselines.}
We adopt camera-controlled video generation models and multi-view videos generation models with point cloud rendering input as baselines for fair comparisons, \ie, ViewCrafter~\cite{viewcrafter}, Reangle-A-Video~\cite{rav}, EX-4D~\cite{ex4d}, and TrajectoryCrafter~\cite{trajectorycrafter}, most of which are recently released state-of-the-art methods. 
See more details for the reproduction of baselines in the Appendix.

\subsection{Comparisons with Other Methods}
\label{sec:comp_sota}

\begin{table*}[t]
\centering
\caption{\textbf{Quantitative ablation results}. I.S.H. means Incremental State Hyperspace. H.G.S. means Hyperspace Guided Sampling.}
\setlength{\tabcolsep}{1.5mm}{
\begin{tabular}{c||cccccccc}
\hline
\cellcolor{gray!10} & \multicolumn{8}{c}{\cellcolor{gray!10}\textbf{MEt3R}$\downarrow$} \\
\cline{2-9}
\cellcolor{gray!10}\multirow{-2}{*}{\textbf{Settings}} & \multicolumn{2}{c}{\cellcolor{gray!10}All Scenes} & \multicolumn{2}{c}{\cellcolor{gray!10}Hard Scenes} & \multicolumn{2}{c}{\cellcolor{gray!10}Internal-View Scenes} & \multicolumn{2}{c}{\cellcolor{gray!10}External-View Scenes} \\
\hline\hline
Ours & 0.1635 & {\color{gray}(+0.00\%)} & 0.2611 & {\color{gray}(+0.00\%)} & 0.1818 & {\color{gray}(+0.00\%)} & 0.1268 & {\color{gray}(+0.00\%)} \\
Ours w/o I.S.H. & 0.1748 & {\color{blue}(-6.91\%)} & 0.2641 & {\color{blue}(-1.15\%)} & 0.1828 & {\color{blue}(-0.55\%)} & 0.1589 & {\color{blue}(-25.32\%)} \\
Ours w/o H.G.S. & 0.1870 & {\color{blue}(-14.37\%)} & 0.3269 & {\color{blue}(-25.20\%)} & 0.2187 & {\color{blue}(-20.30\%)} & 0.1237 & {\color{red}(+2.44\%)} \\
Ours w/o I.S.H., w/o H.G.S. & 0.2136 & {\color{blue}(-30.64\%)} & 0.3421 & {\color{blue}(-31.02\%)} & 0.2291 & {\color{blue}(-26.02\%)} & 0.1826 & {\color{blue}(-44.01\%)} \\
\hline
\end{tabular}
}
\label{tab:comp_abla_1}
\vspace{-1.0em}
\end{table*}

\textbf{Quantitative comparisons with other methods}.
The quantitative experimental results compared to other methods are shown in \cref{tab:comp_sota}.
We can see that our method achieves significant improvements in 3D consistency, \ie, MEt3R~\cite{met3r}, while also outperforming other methods in most metrics of video generation quality, \ie, VBench~\cite{vbench}.
Specifically, our proposed method achieves $15.28\%$ improvement of 3D consistency compared to the state-of-the-art method TrajectoryCrafter~\cite{trajectorycrafter}.
Meanwhile, our method achieves $11.38\%$ and $13.17\%$ improvements of subject consistency and imaging quality, respectively, compared to the prevailing time-synchronized multi-view videos generation method Reangle-A-Video~\cite{rav}.
Furthermore, we also evaluate the performance of all methods under different scene-splits, \ie, hard scenes, internal-view scenes, and external-view scenes, which are reported in the Appendix.

\textbf{Qualitative comparisons with other methods}. 
Qualitative results are visualized in \cref{fig:comp_sotas_1}, showing selected views $[0, 24, 48]$ and timestamps $[0, 24, 48]$ for each scene in a $3 \times 3$ grid for each method.
Horizontal inspection reveals temporal variations within a single view, while vertical alignment assesses 3D consistency across views at each timestamp.
The state-of-the-art method TrajectoryCrafter~\cite{trajectorycrafter} exhibits dynamic content mismatches and 3D inconsistencies across views, attributed to its lack of world state hyperspace guidance during sampling, as discussed in \cref{sec:intro,sec:method}. Other methods~\cite{rav, ex4d} fail to generate plausible content due to fixed mask-gradually-variation patterns or unsuitability for dynamic scenarios~\cite{viewcrafter}.
In contrast, our method constructs an incremental world state hyperspace and employs hyperspace-guided sampling, enabling temporally synchronized multi-view video generation with high visual fidelity and robust 3D consistency. Additional qualitative results with extended views and timestamps are provided in the Appendix.

\subsection{Performance Analysis}
\label{sec:performance_analysis}

\begin{figure}[t]
    \centering
    \includegraphics[width=\columnwidth]{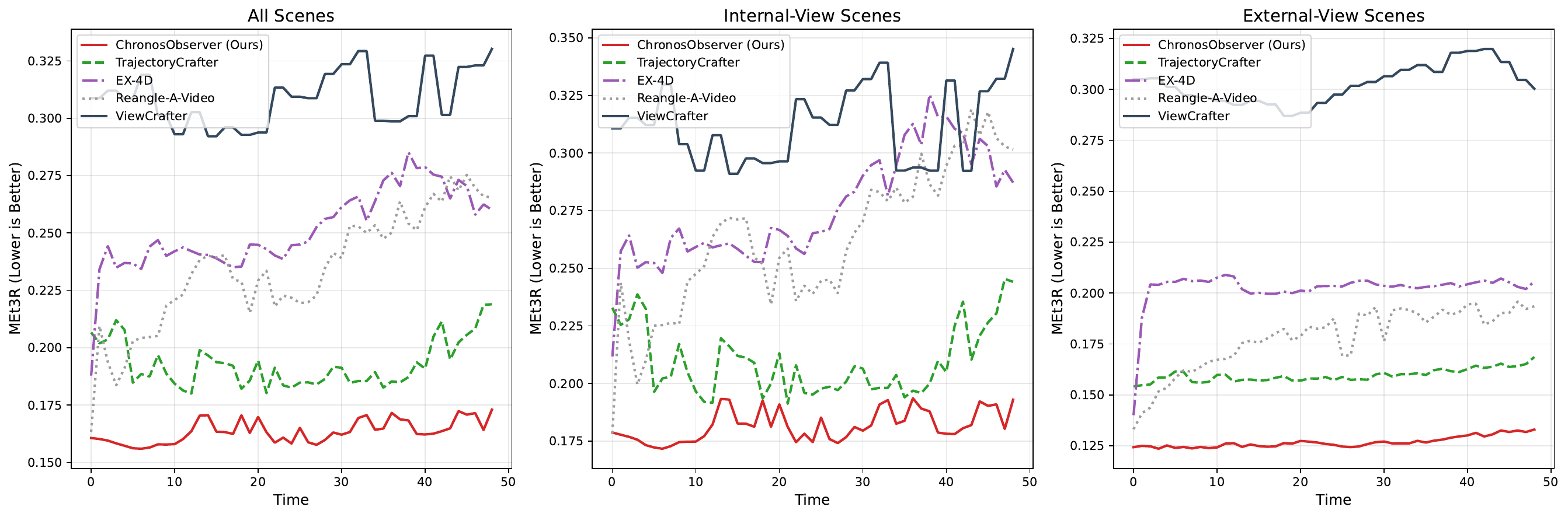}
    \caption{\textbf{Relationship between timestamp and MEt3R}$\downarrow$. Our method demonstrates higher 3D consistency across timestamps compared to other methods.}
    \label{fig:analysis_time_met3r}
    \vspace{-1.0em}
\end{figure}

\textbf{Does ChronosObserver maintain 3D consistency at every timestamp?}
We provide a comprehensive evaluation of 3D consistency across timestamps as shown in \cref{fig:analysis_time_met3r}, where 3D consistency is measured by the MEt3R~\cite{met3r} (lower is better) and averaged by different scenes and views.
The results demonstrate that our method (red line) consistently maintains more stable consistencies over the entire timestamps compared to the prevailing methods.
The MEt3R curve of our proposed method remains persistently lower and exhibits less fluctuation over time than the trend of the other models.
The consistent outperformance demonstrates that our method effectively preserves 3D consistency across timestamps. This robustness highlights the state-of-the-art capability of our method in generating time-synchronized multi-view videos, while prevailing methods exhibit significant fluctuations and 3D inconsistencies across timestamps.

\begin{table*}[t]
\centering
\caption{\textbf{Exploration on static state}, which provides steady improvements in generation quality.}
\setlength{\tabcolsep}{2mm}{
\begin{tabular}{c||c|cccccc}
\hline
\cellcolor{gray!10} & \cellcolor{gray!10}\textbf{3D Consistency} & \multicolumn{6}{c}{\cellcolor{gray!10}\textbf{VBench}} \\
\cline{2-8}
\cellcolor{gray!10}\multirow{-2}{*}{\textbf{Settings}} & \cellcolor{gray!10}MEt3R$\downarrow$ & \cellcolor{gray!10}Subject$\uparrow$ & \cellcolor{gray!10}Background$\uparrow$ & \cellcolor{gray!10}Motion$\uparrow$ & \cellcolor{gray!10}Temporal$\uparrow$ & \cellcolor{gray!10}Aesthetic$\uparrow$ & \cellcolor{gray!10}Imaging$\uparrow$ \\
\hline\hline
Ours w/ Static State & 0.1635 & 0.9365 & 0.9582 & 0.9918 & 0.9834 & 0.5598 & 0.6332 \\
Ours w/o Static State & 0.1787 & 0.9235 & 0.9482 & 0.9885 & 0.9770 & 0.5682 & 0.6144 \\
\hline
\end{tabular}
}
\label{tab:comp_abla_2}
\vspace{-1.0em}
\end{table*}

\begin{figure}[t]
    \centering
    \includegraphics[width=\columnwidth]{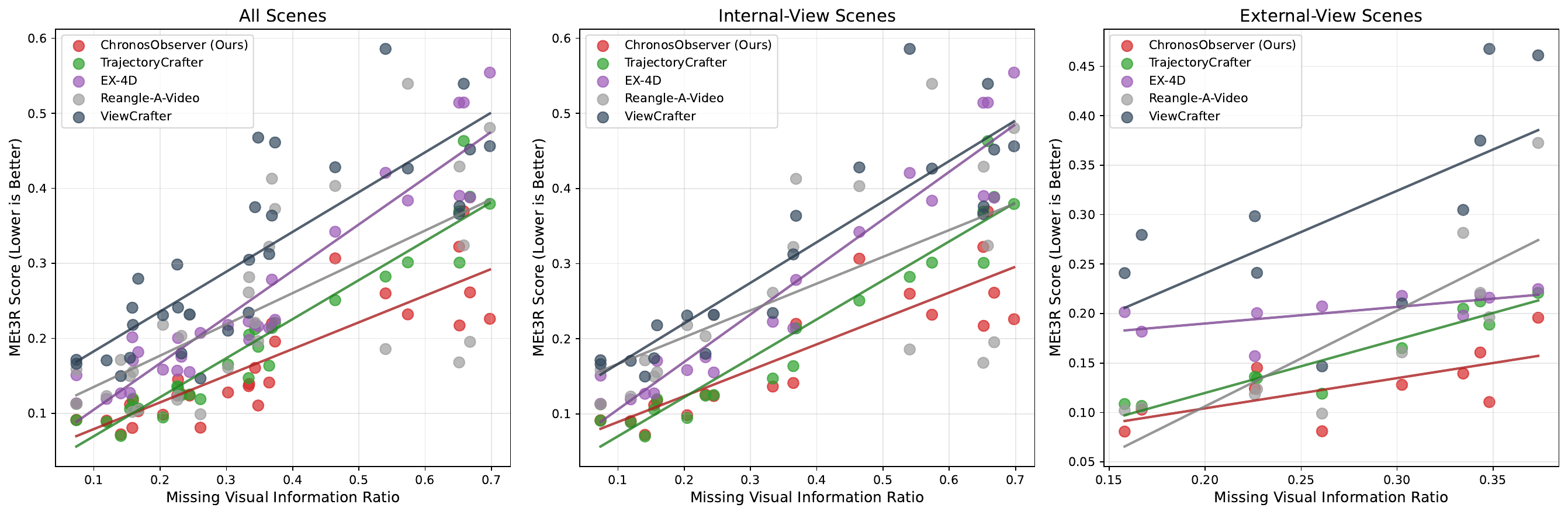}
    \caption{\textbf{Relationship between missing visual information ratio and MEt3R}$\downarrow$. Our method is more effective than other methods in scenarios where more visual information is missing.}
    \label{fig:analysis_mask_met3r}
    \vspace{-0.5em}
\end{figure}

\textbf{Can ChronosObserver handle large areas of missing visual information?}
We visualize the missing visual information ratio and MEt3R~\cite{met3r} (lower is better) for each scene as shown in \cref{fig:analysis_mask_met3r}, where the ratio is computed using the rendering mask of TrajectoryCrafter~\cite{trajectorycrafter}, which can indicate the difficulty of generating visual content for different scenarios in terms of co-visibility between viewpoints.
Our method exhibits a lower overall trend of MEt3R across various scenarios, \ie, better 3D consistency.
Furthermore, the superiority of our proposed method becomes substantially more pronounced as the missing ratio increases. 
Specifically, the MEt3R trendlines of most prevailing methods exhibit a sharper positive slope than ours at high missing ratios, indicating more rapid degradation of other methods in 3D consistency compared to our method.
The significant performance gap under conditions of severe information deficiency demonstrates the robustness of our method to various scenarios and highlights its effectiveness in generating 3D-consistent visual content, which is a key challenge that existing methods struggle with.

\begin{figure}[t]
    \centering
    \includegraphics[width=\columnwidth]{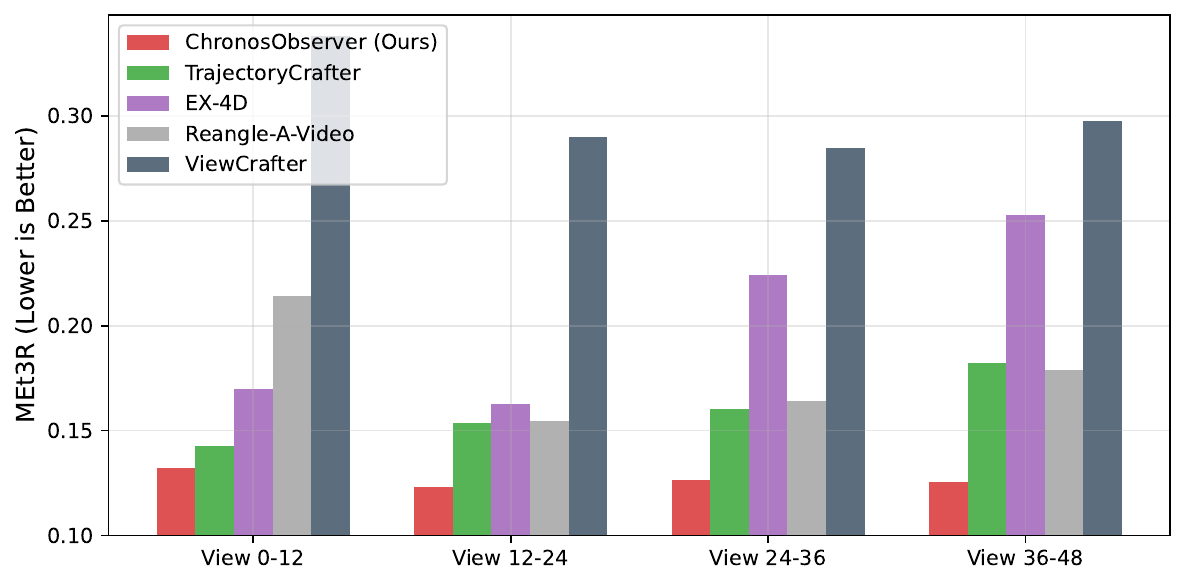}
    \caption{\textbf{Relationship between viewpoints and MEt3R}$\downarrow$ on external-view scenes. Our method achieves better 3D consistency and stability across extrapolated viewpoints.}
    \label{fig:analysis_view_met3r_extrapolate}
    \vspace{-1.0em}
\end{figure}

\textbf{Exploration of ChronosObserver on the extrapolation ability.}
We visualize the pair-wise viewpoints and MEt3R~\cite{met3r} (lower is better) on the external-view scenes as shown in \cref{fig:analysis_view_met3r_extrapolate}.
We can notice that our method achieves an overall better 3D consistency across pair-wise viewpoints compared to other methods.
Furthermore, as the degree of extrapolation increases, \ie, the view number increases, although some methods~\cite{trajectorycrafter, ex4d} show a positive correlation with 3D inconsistency, our method demonstrates better stability and robustness in generating 3D-consistent visual content for extrapolation scenarios.

\textbf{Ablations of ChronosObserver.}
Detailed ablation studies are conducted to evaluate the contributions of Incremental State Hyperspace (I.S.H.) and Hyperspace Guided Sampling (H.G.S.) across different scene splits, \ie, hard scenes, internal-view scenes, and external-view scenes.
The \textbf{quantitative results} of ablations are presented in \cref{tab:comp_abla_1}.
The first row demonstrates the performance of our full model (\textit{Ours}). 
Rows 2–3 validate the individual effectiveness of both I.S.H. and H.G.S., as removing either component results in a measurable performance drop. When both components are ablated simultaneously (last row), a substantial performance decline of $-30.64\%$ is observed on All Scenes. This cumulative degradation indicates that I.S.H. and H.G.S. offer complementary strengths, and their synergistic integration is critical for ensuring robust 3D consistency in multi-view video generation.
\begin{figure}[t]
    \centering
    \includegraphics[width=\columnwidth]{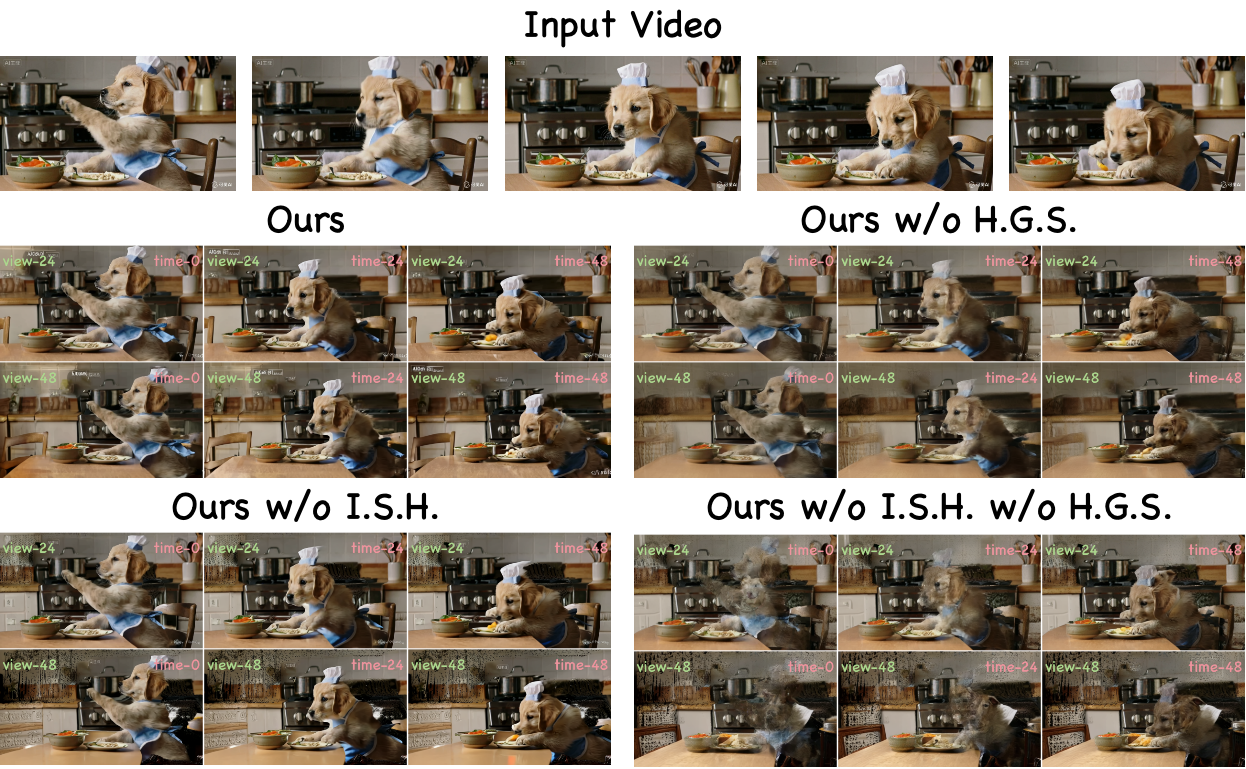}
    \caption{\textbf{Qualitative comparisons of the ablations}. I.S.H. means Incremental State Hyperspace. H.G.S. means Hyperspace Guided Sampling.}
    \label{fig:comp_abla_1}
    \vspace{-1.0em}
\end{figure}
The \textbf{qualitative results} of ablations are presented in \cref{fig:comp_abla_1}.
We can observe that I.S.H. plays a significant role in promoting 3D consistency across different views at the same timestamp and in extrapolating visual content. Furthermore, H.G.S. promotes the guidance of states in the hyperspace for the diffusion sampling process, while removing H.G.S. significantly degrades the visual content generation. Moreover, removing both components simultaneously results in an even more severe degradation in generation quality.

\begin{figure}[t]
    \centering
    \includegraphics[width=\columnwidth]{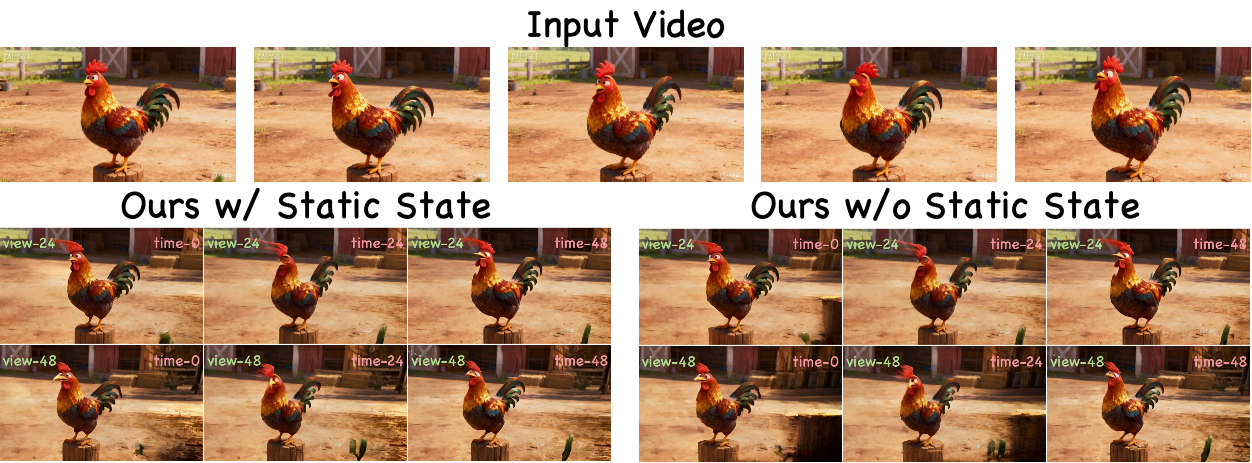}
    \caption{\textbf{Exploration of static state}, which shows the effect of improving the stability of static background generation.}
    \label{fig:comp_abla_2}
    \vspace{-1.0em}
\end{figure}

\textbf{Exploration of Static State.}
Quantitative results of an ablation study investigating the role of base static state $\mathbb{P}_{\Omega}^{\alpha}$ within our pipeline are shown in \cref{tab:comp_abla_2}, which indicates that the incorporation of static state yields an overall and steady improvement of generation quality. 
Qualitative results are provided in \cref{fig:comp_abla_2}. As demonstrated in \cref{sec:method}, the base static state of the dynamic scene is important for generating temporally coherent backgrounds, effectively mitigating structural ambiguities. As shown in the bottom-right corner of the scene presented in \cref{fig:comp_abla_2}, the regions affected by the static state exhibit better static-region visual structural stability compared to the ablated variant.

\begin{table}[t]
\centering
\caption{\textbf{Comparisons on computational time}.}
\setlength{\tabcolsep}{1mm}{
\begin{tabular}{c||cc}
\hline
\cellcolor{gray!10} & \cellcolor{gray!10}3D Consistency & \cellcolor{gray!10}Time/View \\ \cline{2-3}
\cellcolor{gray!10}\multirow{-2}{*}{Methods} & \cellcolor{gray!10}MEt3R$\downarrow$ & \cellcolor{gray!10}Hours$\downarrow$ \\
\hline\hline
Reangle-A-Video~\cite{rav} & 0.2342 & $\approx$0.6756 \\
ChronosObserver (Ours) & 0.1635 & $\approx$0.1676 \\
\hline
\end{tabular}
}
\label{tab:comp_time}
\vspace{-0.5em}
\end{table}
\textbf{Computational Time Cost.}
The computational time cost compared with the state-of-the-art multi-view video generation method Reangle-A-Video~\cite{rav} is shown in \cref{tab:comp_time}. We randomly selected one scene each from internal-view scenes and external-view scenes for testing. Note that Reangle-A-Video~\cite{rav} enables gradient checkpointing due to the limitation of VRAM, and the total time of Reangle-A-Video~\cite{rav} is supplemented by the time of MegaSAM~\cite{megasam} (although relatively short). Then, we divided the total time by the number of target views and reported it as Time/View as shown in \cref{tab:comp_time}, where we further supplement MEt3R for reference.
We can notice that, although our method is training-free, it not only requires less time than the state-of-the-art method of time-synchronized multi-view videos generation, but also achieves better generation results.

\section{Conclusion}
\label{sec:conclusion}

We propose ChronosObserver, a training-free method for time-synchronized multi-view videos generation, which includes World State Hyperspace for incrementally modeling the spatiotemporal constraints of a 4D world scenario, and Hyperspace Guided Sampling for utilizing the constraints within the hyperspace to control the sampling trajectories of the pre-trained camera-controlled video diffusion model, thereby achieving 3D-consistent and high-fidelity results.

The limitation of our method is its dependence on the accuracy of monocular depth estimation and depth completion, a challenge it shares with other methods that rely on monocular depth estimates to achieve frame warping constraints for guiding video diffusion sampling.
{
    \small
    \bibliographystyle{ieeenat_fullname}
    \bibliography{main}
}

\clearpage
\appendix

\section{Additional Experimental Results}
\label{apdx:sec:additional_exp_results}

\subsection{Video and Website}
\label{apdx:subsec:video_website}

We provide additional experimental results in the \url{https://icvteam.github.io/ChronosObserver.html}. We highly recommend watching them.

\subsection{More Quantitative Results}
\label{apdx:subsec:more_quantitative_results}

\textbf{More quantitative results compared to other methods.}
We evaluate the performance of different methods under different scene-splits, \ie, hard scenes, internal-view scenes, and external-view scenes, as shown in the \cref{apdx:tab:quantitative-result}.
It can be observed that in most scenarios, our method achieves the best or second-best video generation quality. More importantly, our approach consistently attains the highest level of 3D consistency across different settings, significantly outperforming other approaches. This demonstrates the robustness and effectiveness of our proposed method.

\begin{figure}[t]
    \centering
    \includegraphics[width=\columnwidth]{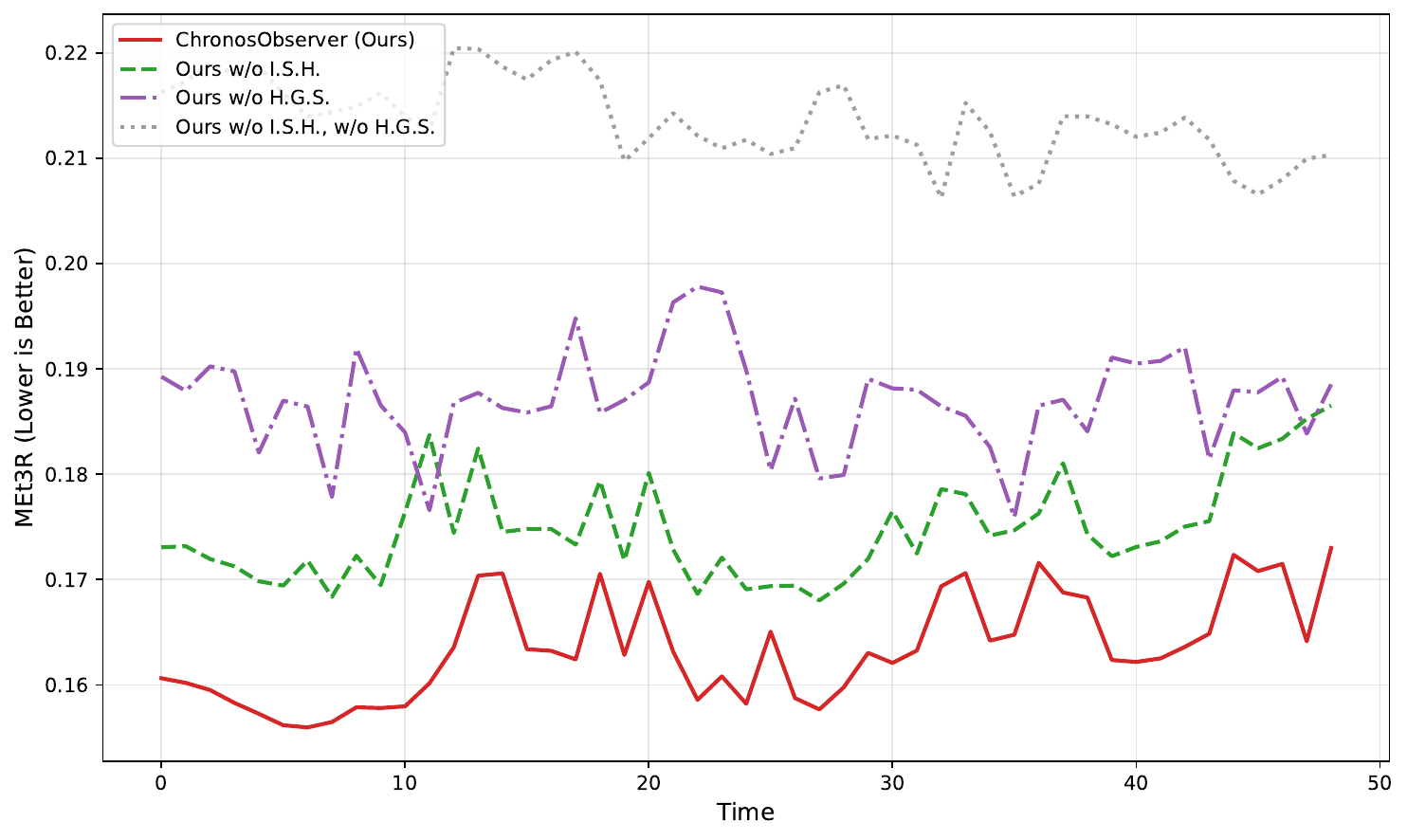}
    \caption{\textbf{Relationship between timestamp and MEt3R}$\downarrow$ for ablations.}
    \label{apdx:fig:analysis_time_abla}
    \vspace{-2em}
\end{figure}

\textbf{More quantitative analysis of ablations.}
Similar to the 3D consistency analysis conducted in the main manuscript across the temporal dimension, we additionally provide an analysis of 3D consistency at different timestamps for the ablations as shown in \cref{apdx:fig:analysis_time_abla}. It can be observed that ablating either I.S.H. (Incremental State Hyperspace) or H.G.S. (Hyperspace Guided Sampling) leads to an apparent degradation in overall 3D consistency across timestamps, while the joint ablation of both components results in further deterioration of 3D consistency across timestamps. This further demonstrates that the enhancement in 3D consistency provided by our proposed method is manifested throughout the entire generated video sequence, thereby underscoring the effectiveness of our proposed method.

\begin{table*}[t]
\centering
\caption{\textbf{More Detailed quantitative comparison results}.}
\setlength{\tabcolsep}{1.5mm}{
\begin{tabular}{cc|cccccc}
\hline
\multirow{2}{*}{\textbf{Metrics}} & \textbf{3D Consistency} & \multicolumn{6}{c}{\textbf{VBench}} \\
\cline{2-8}
& MEt3R$\downarrow$ & Subject$\uparrow$ & Background$\uparrow$ & Motion$\uparrow$ & Temporal$\uparrow$ & Aesthetic$\uparrow$ & Imaging$\uparrow$ \\
\hline
\multicolumn{8}{c}{\textbf{All Scenes}} \\
\hline
ViewCrafter \cite{viewcrafter} & \cellcolor{5th}0.3086 & \cellcolor{3rd}0.9038 & \cellcolor{3rd}0.9403 & \cellcolor{4th}0.9792 & \cellcolor{2nd}0.9829 & \cellcolor{4th}0.4907 & \cellcolor{3rd}0.5909 \\
Reangle-A-Video \cite{rav} & \cellcolor{3rd}0.2342 & \cellcolor{4th}0.8408 & \cellcolor{4th}0.9070 & \cellcolor{5th}0.9788 & \cellcolor{5th}0.9583 & \cellcolor{3rd}0.5303 & \cellcolor{4th}0.5595 \\
EX-4D \cite{ex4d} & \cellcolor{4th}0.2514 & \cellcolor{5th}0.8267 & \cellcolor{5th}0.8935 & \cellcolor{3rd}0.9881 & \cellcolor{4th}0.9769 & \cellcolor{5th}0.4873 & \cellcolor{5th}0.5546 \\
TrajectoryCrafter \cite{trajectorycrafter} & \cellcolor{2nd}0.1930 & \cellcolor{2nd}0.9238 & \cellcolor{2nd}0.9475 & \cellcolor{2nd}0.9886 & \cellcolor{3rd}0.9773 & \cellcolor{1st}0.5660 & \cellcolor{2nd}0.6149 \\
ChronosObserver (Ours) & \cellcolor{1st}0.1635 & \cellcolor{1st}0.9365 & \cellcolor{1st}0.9582 & \cellcolor{1st}0.9918 & \cellcolor{1st}0.9834 & \cellcolor{2nd}0.5598 & \cellcolor{1st}0.6332 \\
\hline
\multicolumn{8}{c}{\textbf{Hard Scenes}} \\
\hline
ViewCrafter \cite{viewcrafter} & \cellcolor{5th}0.4455 & \cellcolor{3rd}0.8371 & \cellcolor{3rd}0.9113 & \cellcolor{4th}0.9668 & \cellcolor{2nd}0.9726 & \cellcolor{4th}0.4528 & \cellcolor{2nd}0.5320 \\
Reangle-A-Video \cite{rav} & \cellcolor{3rd}0.3511 & \cellcolor{5th}0.6655 & \cellcolor{5th}0.8251 & \cellcolor{5th}0.9541 & \cellcolor{5th}0.9136 & \cellcolor{3rd}0.5133 & \cellcolor{4th}0.4733 \\
EX-4D \cite{ex4d} & \cellcolor{4th}0.4011 & \cellcolor{4th}0.6842 &\cellcolor{4th} 0.8399 & \cellcolor{2nd}0.9823 & \cellcolor{3rd}0.9653 & \cellcolor{5th}0.4388 & \cellcolor{5th}0.4392 \\
TrajectoryCrafter \cite{trajectorycrafter} & \cellcolor{2nd}0.3171 & \cellcolor{2nd}0.8557 & \cellcolor{2nd}0.9145 & \cellcolor{3rd}0.9791 & \cellcolor{4th}0.9605 & \cellcolor{1st}0.5441 & \cellcolor{3rd}0.5133 \\
ChronosObserver (Ours) & \cellcolor{1st}0.2611 & \cellcolor{1st}0.8964 & \cellcolor{1st}0.9470 & \cellcolor{1st}0.9891 & \cellcolor{1st}0.9800 & \cellcolor{2nd}0.5213 & \cellcolor{1st}0.5606 \\
\hline
\multicolumn{8}{c}{\textbf{Internal-View Scenes}} \\
\hline
ViewCrafter \cite{viewcrafter} & \cellcolor{5th}0.3117 & \cellcolor{3rd}0.8871 & \cellcolor{3rd}0.9266 & \cellcolor{4th}0.9736 & \cellcolor{2nd}0.9784 & \cellcolor{4th}0.4642 & \cellcolor{3rd}0.5518 \\
Reangle-A-Video \cite{rav} & \cellcolor{3rd}0.2623 & \cellcolor{4th}0.7925 & \cellcolor{4th}0.8778 & \cellcolor{5th}0.9728 & \cellcolor{5th}0.9454 & \cellcolor{2nd}0.5076 & \cellcolor{4th}0.5107 \\
EX-4D \cite{ex4d} & \cellcolor{4th}0.2760 & \cellcolor{5th}0.7875 & \cellcolor{5th}0.8704 & \cellcolor{2nd}0.9876 & \cellcolor{3rd}0.9735 & \cellcolor{5th}0.4326 & \cellcolor{5th}0.4995 \\
TrajectoryCrafter \cite{trajectorycrafter} & \cellcolor{2nd}0.2096 & \cellcolor{2nd}0.9076 & \cellcolor{2nd}0.9317 & \cellcolor{3rd}0.9868 & \cellcolor{4th}0.9725 & \cellcolor{1st}0.5194 & \cellcolor{2nd}0.5850 \\
ChronosObserver (Ours) & \cellcolor{1st}0.1818 & \cellcolor{1st}0.9274 & \cellcolor{1st}0.9486 & \cellcolor{1st}0.9919 & \cellcolor{1st}0.9823 & \cellcolor{3rd}0.5060 & \cellcolor{1st}0.6113 \\
\hline
\multicolumn{8}{c}{\textbf{External-View Scenes}} \\
\hline
ViewCrafter \cite{viewcrafter} & \cellcolor{5th}0.3025 & \cellcolor{4th}0.9373 & \cellcolor{3rd}0.9676 & \cellcolor{4th}0.9904 & \cellcolor{1st}0.9918 & \cellcolor{5th}0.5436 & \cellcolor{3rd}0.6692 \\
Reangle-A-Video \cite{rav} & \cellcolor{3rd}0.1780 & \cellcolor{3rd}0.9375 & \cellcolor{4th}0.9653 & \cellcolor{3rd}0.9907 & \cellcolor{4th}0.9841 & \cellcolor{4th}0.5758 & \cellcolor{5th}0.6570 \\
EX-4D \cite{ex4d} & \cellcolor{4th}0.2023 & \cellcolor{5th}0.9052 & \cellcolor{5th}0.9397 & \cellcolor{5th}0.9891 & \cellcolor{5th}0.9838 & \cellcolor{3rd}0.5966 & \cellcolor{4th}0.6647 \\
TrajectoryCrafter \cite{trajectorycrafter} & \cellcolor{2nd}0.1597 & \cellcolor{1st}0.9562 & \cellcolor{1st}0.9791 & \cellcolor{1st}0.9921 & \cellcolor{2nd}0.9868 & \cellcolor{2nd}0.6591 & \cellcolor{2nd}0.6747 \\
ChronosObserver (Ours) & \cellcolor{1st}0.1268 & \cellcolor{2nd}0.9547 & \cellcolor{2nd}0.9772 & \cellcolor{2nd}0.9916 & \cellcolor{3rd}0.9856 & \cellcolor{1st}0.6674 & \cellcolor{1st}0.6770 \\
\hline
\end{tabular}
}
\label{apdx:tab:quantitative-result}
\end{table*}

\subsection{More Qualitative Results}
\label{apdx:subsec:more_qualitative_results}

\begin{figure*}[t]
    \centering
    \includegraphics[width=\textwidth]{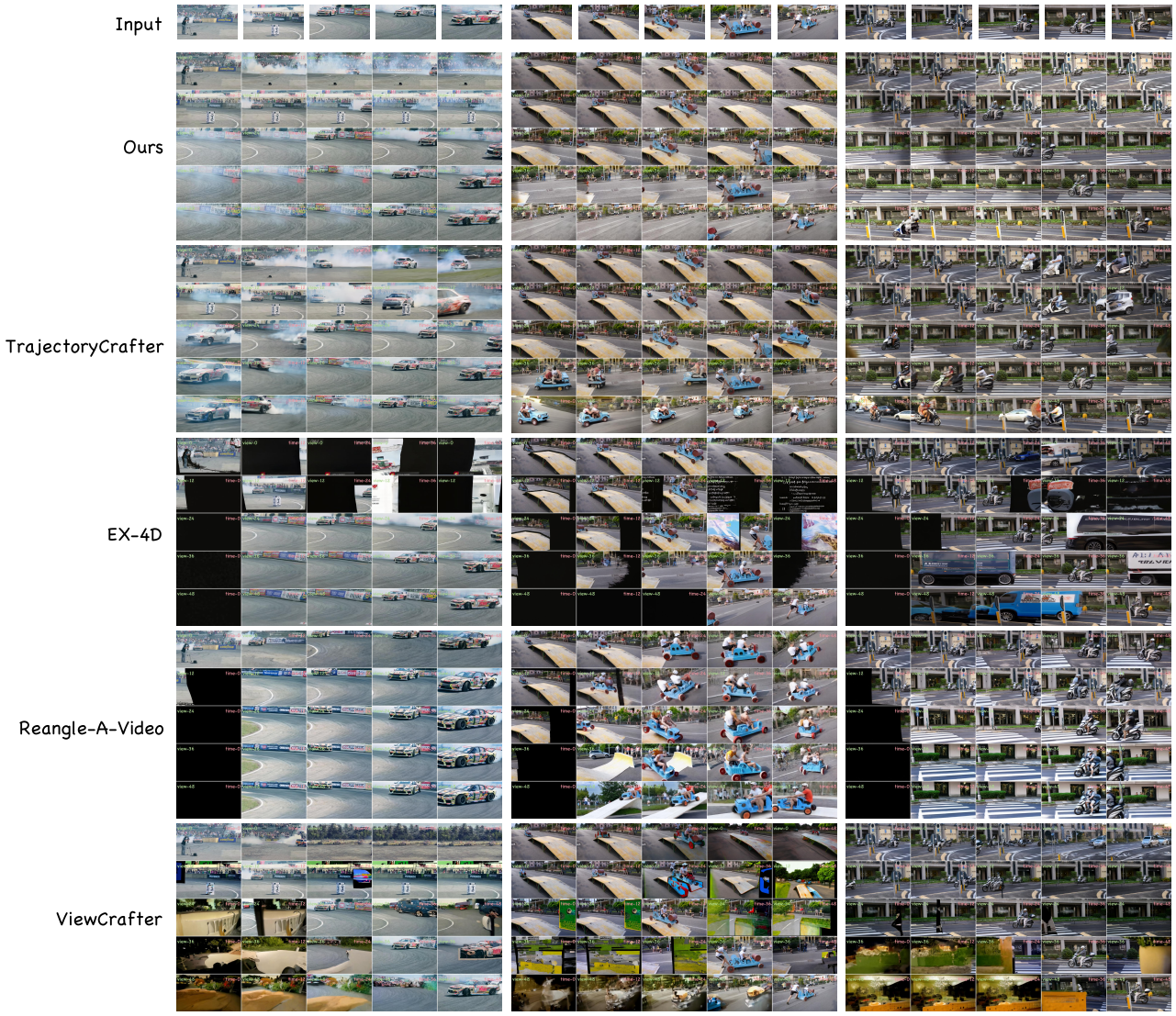}
    \caption{\textbf{Qualitative comparisons with other methods of the internal-view scenes}.}
    \label{apdx:fig:comp_sotas_2}
\end{figure*}

\begin{figure*}[t]
    \centering
    \includegraphics[width=\textwidth]{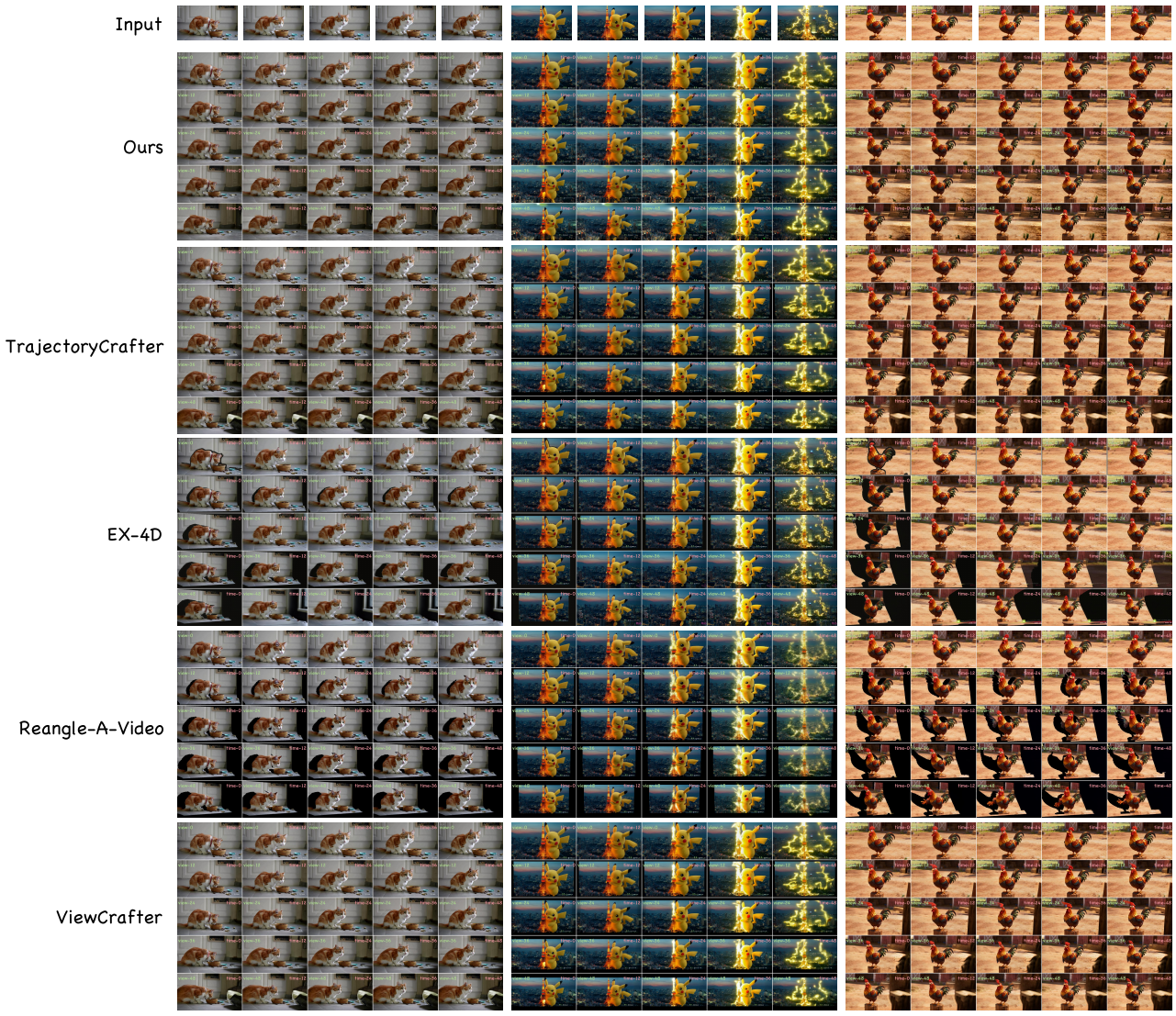}
    \caption{\textbf{Qualitative comparisons with other methods of the external-view scenes}.}
    \label{apdx:fig:comp_sotas_3}
\end{figure*}

\textbf{More qualitative results compared to other methods.}
We provide more qualitative results compared to other methods shown in \cref{apdx:fig:comp_sotas_2} for the internal-view scenes and \cref{apdx:fig:comp_sotas_3} for the external-view scenes.
More qualitative results of comparisons with other methods are provided in the \textbf{demo video} and the \textbf{local static website} of the Supplementary Materials.
Similar to the main manuscript, qualitative results are visualized showing selected views $[0, 12, 24, 36, 48]$ and timestamps $[0, 12, 24, 36, 48]$ for each scene in a $5 \times 5$ grid for each method.
Horizontal inspection reveals temporal variations within a single view, while vertical alignment assesses 3D consistency across views at each timestamp.
The state-of-the-art method TrajectoryCrafter~\cite{trajectorycrafter} exhibits noticeable dynamic content mismatches and 3D inconsistencies across different viewpoints. These limitations stem primarily from the absence of world state hyperspace guidance during the sampling process, as analyzed in the main manuscript. Without such guidance, the model fails to maintain a coherent world scene representation over time, leading to visible artifacts and geometric misalignments across different viewpoints.
Other baseline approaches, including Reangle-A-Video~\cite{rav} and EX-4D~\cite{ex4d}, also struggle to produce convincing multi-view dynamic content. They rely on fixed mask-gradual-variation patterns that lack adaptability to complex scene dynamics. while ViewCrafter~\cite{viewcrafter} is fundamentally designed for static scenes, making it unsuitable for generating temporally coherent video content in dynamic settings.
In contrast, our method introduces an incremental world state hyperspace construction process, coupled with hyperspace guided sampling, which ensures that the generated time-synchronized multi-view videos remain consistent across all views at each timestamp, thereby achieving time-synchronized multi-view videos generation with high visual fidelity and significantly improved 3D consistency compared to existing methods.

\begin{figure*}[t]
    \centering
    \includegraphics[width=0.8\textwidth]{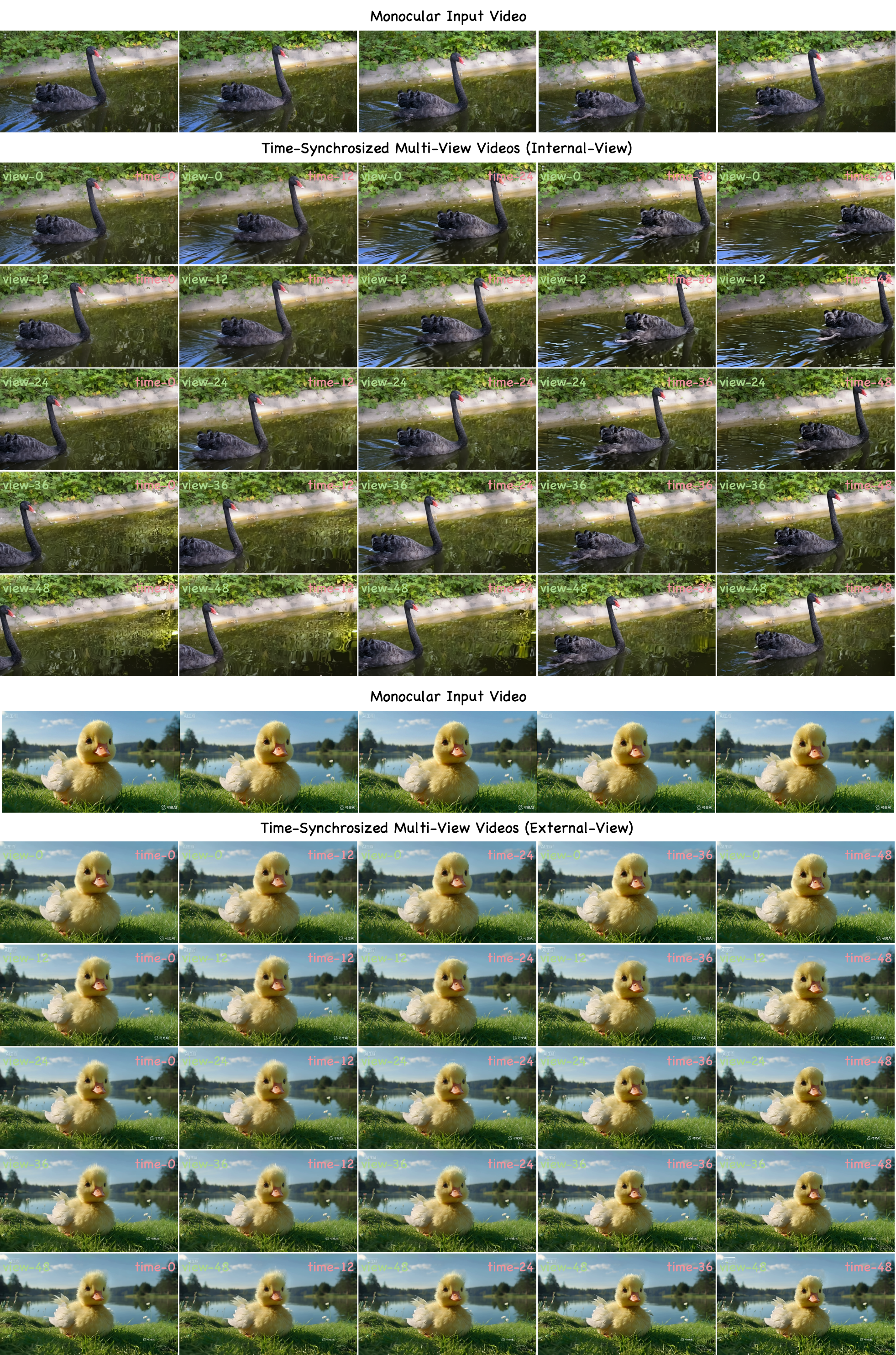}
    \caption{\textbf{Generated time-synchrosized multi-view videos from our method}.}
    \label{apdx:fig:ours_results_1}
\end{figure*}

\begin{figure*}[t]
    \centering
    \includegraphics[width=0.8\textwidth]{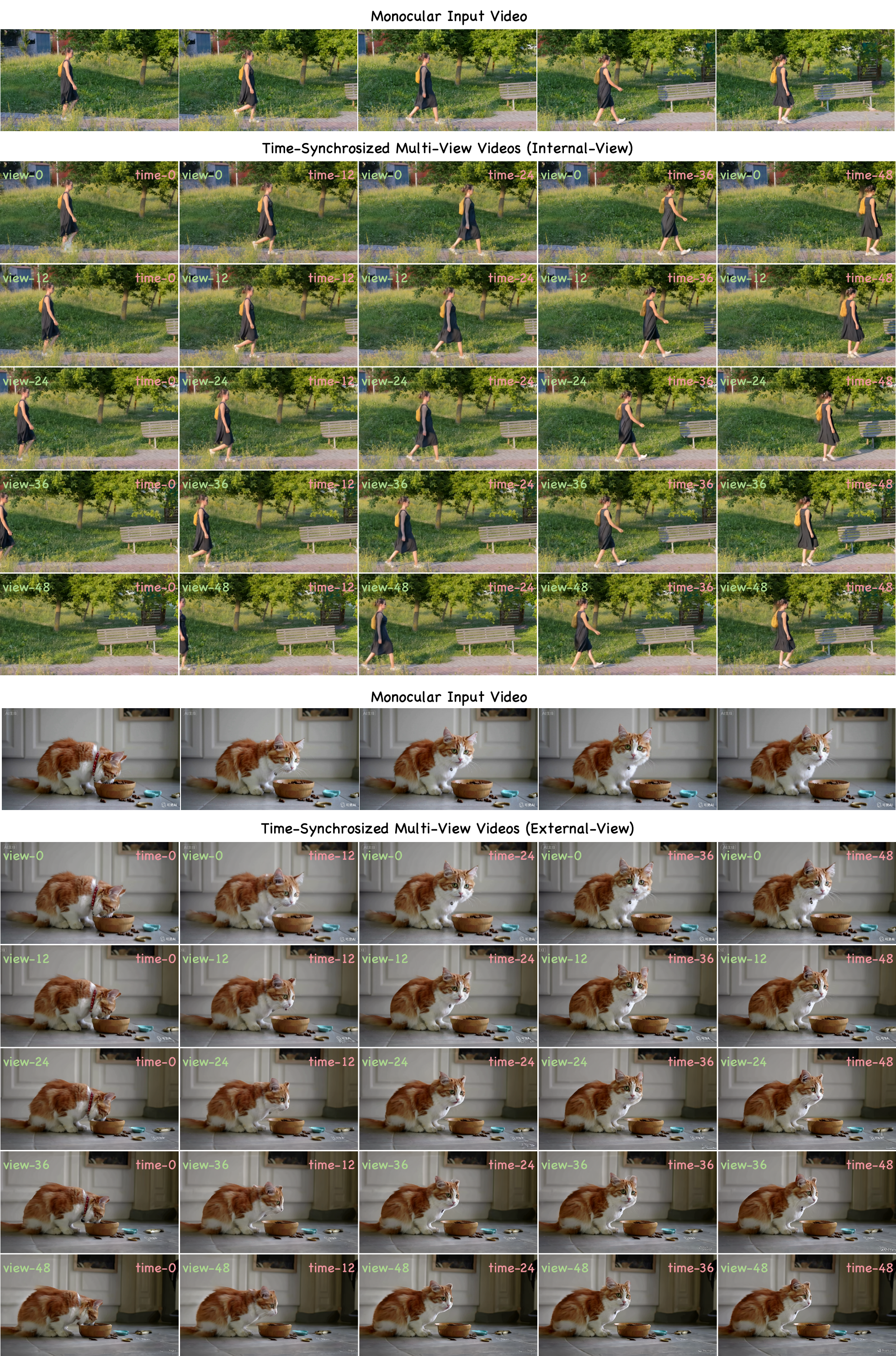}
    \caption{\textbf{Generated time-synchrosized multi-view videos from our method}.}
    \label{apdx:fig:ours_results_2}
\end{figure*}

\textbf{More qualitative results of our method.}
We provide the visualization of the generated time-synchronized multi-view videos from our proposed ChronosObserver in \cref{apdx:fig:ours_results_1} and \cref{apdx:fig:ours_results_2}.
More qualitative results of the generated time-synchronized multi-view videos from our proposed ChronosObserver are provided in the \textbf{demo video} and the \textbf{local static website} of the Supplementary Materials.

\section{Additional Details}
\label{apdx:sec:additional_details}

\subsection{Dataset Details}
\label{apdx:subsec:dataset_details}

We evaluate our method on a collected dataset of 30 monocular video sequences, comprising 20 real-world videos from DAVIS~\cite{davis} and 10 generated videos by the Kling model~\cite{kling}. Each video is preprocessed to contain 49 frames at a resolution of $384\times672$, with longer original videos being truncated and frames extracted accordingly. 
The dataset is fundamentally divided into two categories based on camera motion characteristics. 
\textbf{Interval-View Scenes} refer to real-world videos from DAVIS~\cite{davis} with relatively wide camera trajectories, for which inference is performed using the fixed internal perspective of the original viewpoints.
\textbf{External-View Scenes} consist of generated videos from Kling~\cite{kling} with relatively static camera trajectories, where extrapolated external perspectives are provided for inference. Note that we directly utilize the generated videos from the Kling~\cite{kling} website.
Additionally, to evaluate performance under challenging conditions, a subset termed \textbf{Hard Scenes} is split by selecting the top-10 cases ranked by the MEt3R~\cite{met3r} scores obtained from the state-of-the-art method TrajectoryCrafter~\cite{trajectorycrafter}.

\subsection{Implementation Details}
\label{apdx:subsec:implementation_details}

Our implementation utilizes TrajectoryCrafter~\cite{trajectorycrafter} served as the foundational camera-controlled video generation model. The diffusion sampling process is configured with 30 denoising steps and the CFG scale of 6.0. For both Interval-View and External-View scenes, we uniformly sample five target viewpoints corresponding to indices [0, 12, 24, 36, 48] from the camera trajectory with 49 viewpoints to evaluate multi-view 3D consistency. 
Our method also follows the order of inferring based on the indices of viewpoints.
Since there are currently few depth completion tools for videos, we utilize a recently released image depth completion tool called Prior Depth Anything (PDA)~\cite{pda} in conjunction with flow-based inter-frame smoothing to implement the incremental construction of world states based on the generated videos. The pattern of PDA is set to 500.
Besides, the initial incremental state related to the auxiliary reference $Q_0$ is obtained through the re-running of MegaSAM~\cite{megasam}, while other incremental states utilize the PDA~\cite{pda}.
TSDF~\cite{tsdf} is implemented using Open3D.
The framework is implemented in PyTorch and runs on CUDA 12.1, with experiments conducted on an NVIDIA RTX 4090 GPU with 48GB of VRAM. 

\subsection{Experimental Details}
\label{apdx:subsec:experimental_details}
We provide additional details regarding the experimental setup for the ablations. When performing H.G.S. ablation, we directly project the states in the hyperspace independently to obtain the corresponding rendered video, which is then used as the condition for sampling. The feature fusion weights are calculated as scalars based on mask ratios. Furthermore, for the ablation of the Static State, we supplement the weights of the denoised output of the base dynamic state due to the elimination of the base static state, for the residual uncovered areas.

\subsection{Metrics Details}
\label{apdx:subsec:metrics_details}

\textbf{VBench.}
We utilize VBench~\cite{vbench} to evaluate the quality of generated videos, including Subject Consistency, Background Consistency, Motion Smoothness, Temporal Flickering, Aesthetic Quality, and Imaging Quality. Dynamic Degree is excluded because this metric is less effective at reflecting the quality of video generation for multi-view videos generation. Since we select 5 viewpoints, each scenario includes 5 evaluation results of VBench~\cite{vbench}. We calculate the average of these metrics for each scenario, and then average them across different scenarios to obtain the final evaluation result.

\textbf{MEt3R.}
We utilize MEt3R~\cite{met3r} to evaluate the 3D consistency of generated videos from different viewpoints across timestamps.
Specifically, for each scenario, we compute the MEt3R metric between every consecutive pair of viewpoints, \ie, 0–12, 12–24, 24–36, and 36–48, at each timestamp. The MEt3R result for a given scenario is then obtained by averaging these values across both the temporal and viewpoint dimensions.
So, it also enables additional performance analysis along the temporal and viewpoint dimensions independently, as demonstrated in the main manuscript.
The overall MEt3R evaluation result is derived by averaging the results across different scenarios.

\subsection{Reproduction of Other Methods}
\label{apdx:subsec:reproduction_details}

In this section, we provide reproduction details of different methods. In summary, we make necessary adjustments during the reproduction process, such as aligning the coordinates of poses. Furthermore, the specific details of the different methods are described below.

\textbf{ViewCrafter.}
Since ViewCrafter~\cite{viewcrafter} supports the resolution of $576 \times 1024$ with $25$ frames, we first resize the rendered image to the corresponding resolution and perform frame extraction after warping. After obtaining the generated result, we then resize it to the original resolution and pad it to 49 frames for subsequent evaluation. Note that although ViewCrafter~\cite{viewcrafter} forces the first frame of the rendered result to be the original image, namely, the first pose of the camera trajectory remains consistent with the original frame, we deactivate this manual operation due to the task settings for multi-view videos generation.

\textbf{Reangle-A-Video.}
Since Reangle-A-Video~\cite{rav} supports the resolution of $480 \times 720$, and the running process will automatically resize images to the corresponding resolution, we resize it to the original resolution after obtaining the generated result for subsequent evaluation.
Note that since Reangle-A-Video~\cite{rav} does not release the code of static-view transport generation, we make necessary adjustments to its existing dynamic camera control generation code to ensure it runs successfully.

\textbf{EX-4D.}
Similar to ViewCrafter~\cite{viewcrafter}, EX-4D~\cite{ex4d} also forces the first frame of the rendered result to be the original image from the input monocular video in its running process, namely, the first pose of the camera trajectory remains consistent with the original first frame. We deactivate this manual operation due to the task settings for multi-view videos generation. 

\textbf{TrajectoryCrafter.}
We set the CFG of TrajectoryCrafter~\cite{trajectorycrafter} to 6.0 and the sampling steps to 30, which is the same as our method. Furthermore, we make additional improvements to its pose construction and implement PyTorch-based image warping, all within the same settings as our method to ensure a fair comparison. 

% {
%     \small
%     \bibliographystyle{ieeenat_fullname}
%     \bibliography{main}
% }

\end{document}